\documentclass[11pt]{book}
\usepackage{naacl}
\usepackage[hyphens]{url}

\newcommand{\thesisTitle}{UW CSE Senior Thesis}
\newcommand{\authorName}{Javon Hickmon}

\usepackage{amsfonts}
\usepackage{multirow,tabularx}
\usepackage{listings}
\usepackage{makecell}
\usepackage{amssymb}
\usepackage{graphicx}
\usepackage{amsmath}
\usepackage{float}
\usepackage{fullpage}
\usepackage{mathrsfs}
\usepackage{subcaption}
\usepackage{hyperref}% http://ctan.org/pkg/hyperref
\hypersetup{
    colorlinks=true,
    linkcolor=blue,
    filecolor=black,      
    urlcolor=blue,
    citecolor=blue
}
\usepackage{mdwlist}
\usepackage{xspace}
\usepackage{setspace}
\usepackage{times} % Set the typeface to Times Roman
\usepackage[scaled]{helvet} % ss
\usepackage{courier} % tt
\normalfont
\usepackage[T1]{fontenc}

\usepackage{changepage}
\usepackage[labelfont=bf]{caption} % boldface caption title for floats
\usepackage[square]{natbib}
\usepackage{enumitem}
\usepackage{natbib}
\usepackage{array,multirow}
\usepackage{fixltx2e}
\usepackage{booktabs}
\usepackage{bbm}
\usepackage{soul}
\usepackage{relsize}
\usepackage{eso-pic}
\usepackage{xspace}
\usepackage{epsfig}
\usepackage{graphicx}
\usepackage{amsmath}
\usepackage{amssymb}
\usepackage{float}
\usepackage{multirow}
\usepackage{rotating}
\usepackage{balance}
\usepackage{wrapfig}
\usepackage{enumerate}
\usepackage{caption}
\usepackage{framed}
\usepackage{enumitem}
\usepackage{multirow}
\usepackage{graphicx}
\usepackage{color}
\usepackage{fixltx2e}
\usepackage{caption}
\usepackage{subcaption}
\usepackage{tikz}
\usepackage{mathtools}
\usepackage{pifont}
\usepackage{scrextend}
\usepackage{sidecap}
\usepackage{graphicx}

\usepackage{pdfpages}
\usepackage{algorithm}
\usepackage{algorithmic}
\usepackage{xcolor,colortbl}
\usepackage{cleveref}
\crefname{figure}{Figure}{Figures}
\crefname{table}{Table}{Tables}
\crefname{section}{Section}{Sections}
\crefname{chapter}{Chapter}{Chapters}
\crefname{appendix}{Appendix}{Appendixes}
\usepackage{multirow}
\usepackage{makecell}
\begin{document}

\includepdf[pages=1]{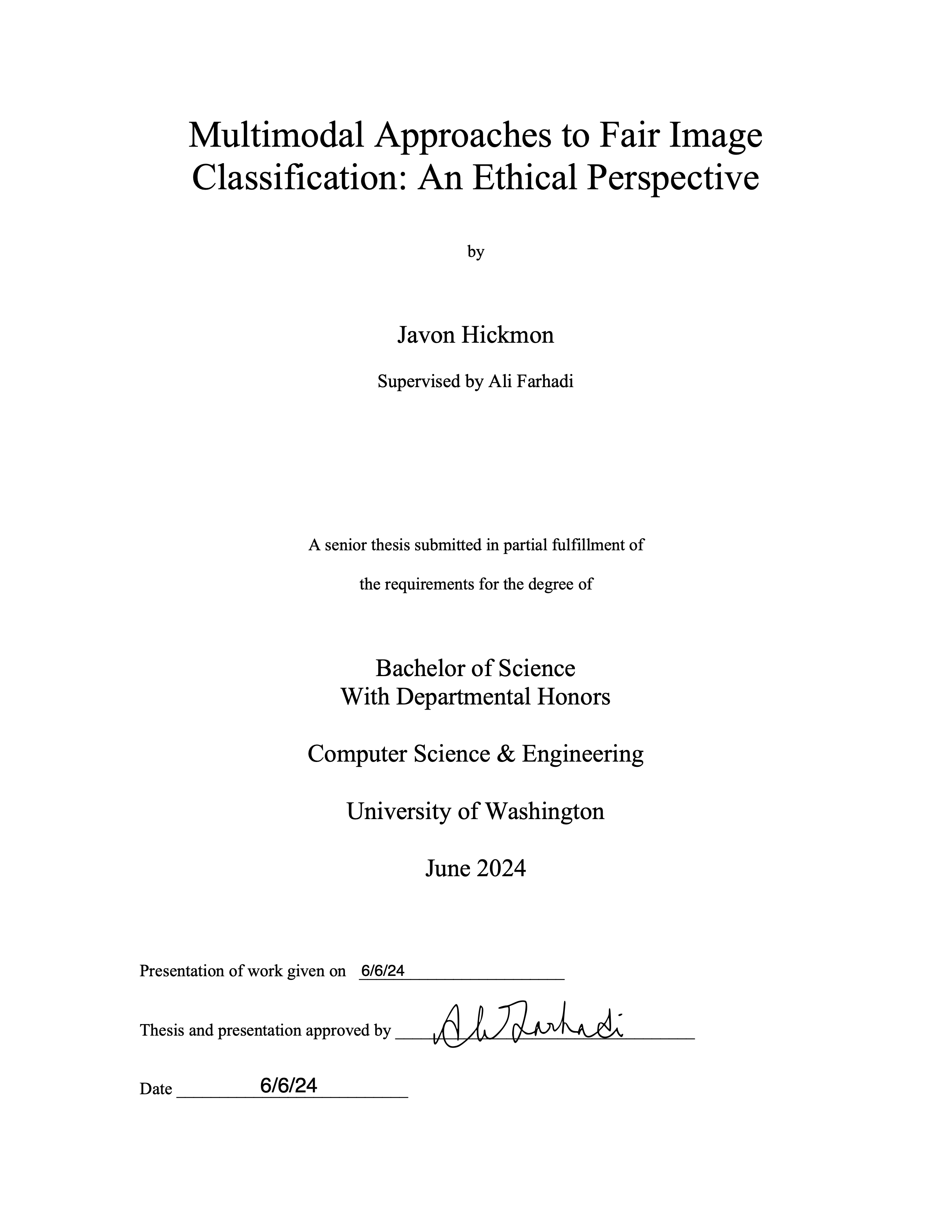}

\doublespacing

%% Replace all of the orange portions with your personal info

\thispagestyle{empty}
\begin{centering}
\vspace{1in}
University of Washington \\
\vspace*{1.\baselineskip}
{\bf Abstract}\\
\vspace*{1\baselineskip}

{\thesisTitle}\\ %self-explanatory
\vspace*{1.\baselineskip}
{\authorName} \\ %self-explanatory
\vspace*{1.\baselineskip}
\end{centering}
% \vspace*{\baselineskip}
In the rapidly advancing field of artificial intelligence, machine perception is becoming paramount to achieving increased performance. Image classification systems are becoming increasingly integral to various applications, ranging from medical diagnostics to image generation; however, these systems often exhibit harmful biases that can lead to unfair and discriminatory outcomes. Machine Learning (ML) systems that depend on a single data modality—i.e., only images or only text—can exaggerate hidden biases present in the training data, if the data is not carefully balanced and filtered. Even so, these models can still harm underrepresented populations when used in improper contexts, such as when government agencies reinforce racial bias using predictive policing. This thesis explores the intersection of technology and ethics in the development of fair image classification models. Specifically I focus on improving \textbf{fairness} and methods of using \textbf{multiple modalities} to combat harmful demographic bias. Integrating multimodal approaches, which combine visual data with additional modalities such as text and metadata, allows this work to enhance the fairness and accuracy of image classification systems. The study critically examines existing biases in image datasets and classification algorithms, proposes innovative methods for mitigating these biases, and evaluates the ethical implications of deploying such systems in real-world scenarios. Through comprehensive experimentation and analysis, the thesis demonstrates how multimodal techniques can contribute to more equitable and ethical AI solutions, ultimately advocating for responsible AI practices that prioritize fairness.

\chapter*{Acknowledgements}
%% self explanatory:
I would like to my mentor Sarah Pratt for advising and assisting with this project. Despite having her own research as a Ph.D. student, she has allowed me to help with her ideas and motivated me to explore my own research questions, leading to the development of this thesis. I would not have been able to explore this work in such depth without her help.\\
\\
I would also like to thank Dr. Orit Peleg for her mentorship and guidance in developing my skills as an independent researcher, Dr. Hyo Lee for her advice in refining my research questions, and the University of Washington McNair team for providing a myriad of research opportunities along with unwavering support.\\
\\
Finally I would like to thank my family. Mom, Ariea, Aunt Sharron, and Uncle Duncan, thank you all so much for your support and encouragement throughout this journey, and for motivating me to create an impact on the world. This work wouldn't exist without you.

\tableofcontents{}
\listoffigures
\listoftables

\chapter {Introduction}
\label{chapter:intro}
\section{Overview}

The rapid development and deployment of artificial intelligence (AI) systems have revolutionized numerous sectors, leading to significant advancements in automation, efficiency, and decision-making processes. Among these AI systems, image classification has emerged as a crucial technology with applications spanning from medical diagnostics, to image generation. Despite its impact, the use of image classification algorithms has also raised serious ethical concerns, particularly regarding fairness and bias. These concerns have been underscored by numerous real-world examples where image classification systems have exhibited discriminatory behavior, leading to adverse outcomes for certain demographic groups.

\section{Background}
One of the most prominent examples of the misuse of image classification technology is in the realm of law enforcement and security. Facial recognition systems, which rely heavily on image classification algorithms, have been widely adopted by law enforcement agencies around the world. However, studies have shown that these systems often perform poorly on individuals with darker skin tones, women, and younger people. A study conducted by the National Institute of Standards and Technology (NIST) found that the error rates for facial recognition systems were significantly higher for people of African and Asian descent compared to those of European descent \citep{NIST_2020}. This disparity has led to wrongful arrests and detentions, as evidenced by the case of Robert Williams, an African American man who was wrongfully arrested in Detroit due to a false match generated by a facial recognition system.\\
\\
In the healthcare sector, image classification algorithms are increasingly used to aid in medical diagnostics, such as detecting tumors in radiological images or identifying skin diseases. However, these systems can also exhibit biases that affect their diagnostic accuracy. For example, multiple studies have found that dermatology AI systems repeatedly performed worse on images of skin conditions from darker-skinned patients compared to those from lighter-skinned patients \citep{marsden2023effectiveness, jones2022artificial}. This discrepancy can result in misdiagnoses or delayed treatment for patients with darker skin, exacerbating existing health disparities. This is reinforced by another study published by the The Lancet Digital Health, that systematically reviewed 21 open access datasets and discovered darker skin types were significantly underrepresented \citep{wen2022characteristics}. This form of under-representation within datasets, is a common theme that extends to the majority of image classification systems, which we will explore further in later chapters.\\
\\
Another concerning example is the use of image classification in social media and online platforms. These platforms employ image recognition algorithms to moderate content, tag images, and personalize user experiences. However, biases in these algorithms can lead to the exclusion or misrepresentation of certain groups. In one notable incident, an image classification system used by Google Photos misidentified photos of Black individuals as "gorillas," highlighting the severe consequences of biased AI systems and the need for more inclusive training data and algorithms.\\
\\
\section{Motivation}
These real-world examples are not even a fraction of the many incidents regarding image classification, and they illustrate the critical need for developing fair techniques and systems. The ethical implications of deploying biased AI systems are profound, affecting individual lives, societal trust in technology, and the broader quest for social justice. Addressing these issues \textbf{requires} a multifaceted approach that combines technological innovation with ethical considerations.\\
\\
This thesis is motivated by the urgent need to enhance the fairness and reduce the harmful bias of image classification systems. The primary goal of this research is to explore how multimodal approaches can be leveraged to mitigate biases in image classification. Multimodal approaches involve the integration of multiple types of data—such as visual data, text, and more—to provide a more comprehensive and context-aware understanding of the input images. Within this work I focus on purely visual and textual data; however, future work will likely explore additional modalities. By incorporating diverse data sources, these approaches can potentially address the limitations of traditional image-only classification systems and improve their performance across different demographic groups.\\
\\
The motivation for this work is rooted in the recognition that fair image classification is not merely a technical challenge but a \textbf{societal imperative}. As AI continues to permeate various aspects of our lives, it is crucial to ensure that these systems are designed and implemented in ways that promote equity and justice. By exploring multimodal approaches to image classification and their ethical implications, this thesis aspires to pave the way for more equitable solutions that benefit all members of society.

\section{Challenges}
With the primary focus being on fair and ethical image classification, there are a number of challenges and we acknowledge that in practice we likely will not be able to create a truly fair system. This is why we intentionally utilize language such as \textit{reduce harmful bias}, because our goal is to mitigate the negative impact of these systems while still maintaining and improving their efficacy. We will examine the potential benefits and risks associated with multimodal classification techniques, considering factors such as privacy, transparency, and accountability. By providing a balanced perspective on the ethical dimensions of AI, this work contributes to the development of responsible AI practices that prioritize fairness.

\section{Approach}
Our approach to mitigate these harms is two-fold. First, we develop a technique called MuSE (Multimodal Synthetic Embeddings) for image classification, which we use to augment the embedding space of pre-trained multimodal models at inference time. MuSE focuses on improving the generalizability of multimodal models being used for classification. This approach means that it can improve performance on a range of standard classification benchmarks with little to no overhead. We leverage the significant benefits of MuSE within our second step.\\
\\
Diverse Demographic Data Generation (D3G) is a technique developed to reduce demographic bias within multimodal models at inference time. Since this technique leverages the benefits of MuSE, it is completely training-free and can be quickly utilized with any multimodal model being used for classification. This is important because this technique works to boost accuracy and reduce demographic bias at the same time.

\subsection{Outline}
The structure of this thesis is as follows:
\begin{itemize}
  \item We conduct a comprehensive review of existing literature on image classification, multimodal model ensembling, and fairness in Deep Learning. This review will highlight the key challenges and gaps in current research and practice.
  \item We propose MuSE, a technique designed to improve the accuracy and generalizability of multimodal models at inference time. To understand the impact of MuSE, we quantify its performance on a range of benchmarks and conduct an analysis of the improvements.
  \item We propose D3G, a similar technique to MuSE that focuses on reducing demographic bias within multimodal models. This technique will be rigorously tested through experiments designed to expose demographic bias within the model. The results of these experiments will then be analyzed to assess the effectiveness of the proposed methods in reducing bias and improving classification accuracy.
  \item Finally, we will discuss the socio-ethical implications of such techniques if deployed within real-world scenarios. This includes a discussion on where/when this technique should be utilized and, more importantly, where/when it should not be used.
\end{itemize}

\chapter{MuSE}
\label{chapter:muse}
\section{Overview}
Artificial intelligence has made significant progress in image classification, an essential task for machine perception to achieve human-level image understanding. Despite this importance, many issues can arise from traditional multimodal image classification methods; however, two are particularly challenging. First, fine-grained image classification is unachievable if the model is too small and underfits the data. Second, if the model wasn’t trained on enough high-quality data or a modality lacks enough representation of a class, the model may contain inherent bias. We propose Multimodal Synthetic Embeddings as a framework to reduce the impact of these issues on pre-trained models. By leveraging open-vocabulary models, We aim to generate vision-language descriptions that emphasize the differences between similar classes, improving fine-grained classification accuracy without any additional training or fine-tuning.

\section{Introduction}
Multimodal systems have been a promising new paradigm in the field in the field of machine learning. Utilizing multiple modalities has resulted in machine learning models that are more accurate and generalizable on a broad range of tasks, including sentiment analysis \cite{wang2020transmodality} and cross-modal retrieval \cite{kim2021vilt}. Despite this performance, Issues such as data redundancy \cite{dittrich2000data}, noise \cite{xiong2006enhancing}, and class imbalance \cite{longadge2013class} are just a few of the many issues that arise from collecting large amounts of data to train on. Despite this need for high-quality data, open-vocabulary models achieve high classification accuracy across many datasets without labeled training data. To accomplish this, these models leverage the massive amounts of image text pairs available online by learning to associate the images with their correct caption, leading to greater flexibility during inference \cite{pratt2023}. These models understand the underlying semantics and can generalize to unseen data by learning from this information.

\section{Related Work}

\textbf{\textit{What does a platypus look like? Generating customized prompts for zero-shot image classification}} \citep{pratt2023} introduces CuPL (Customized Prompts via Language models), the foundational work for this project. In this paper, generating descriptive prompts for each ImageNet class directly improved the resulting zero-shot accuracy. The work solidifies the idea that prompt generation from a larger model trained on more high-quality data can lead to knowledge transfer, helping to emphasize the differences between classes. Although the attention of the multimodal model is shifted when given generative CuPL descriptions, the model can still fail to distinguish between similar classes, as shown in the paper.

\begin{figure}[h]
    \begin{center}
        \includegraphics[width=14cm]{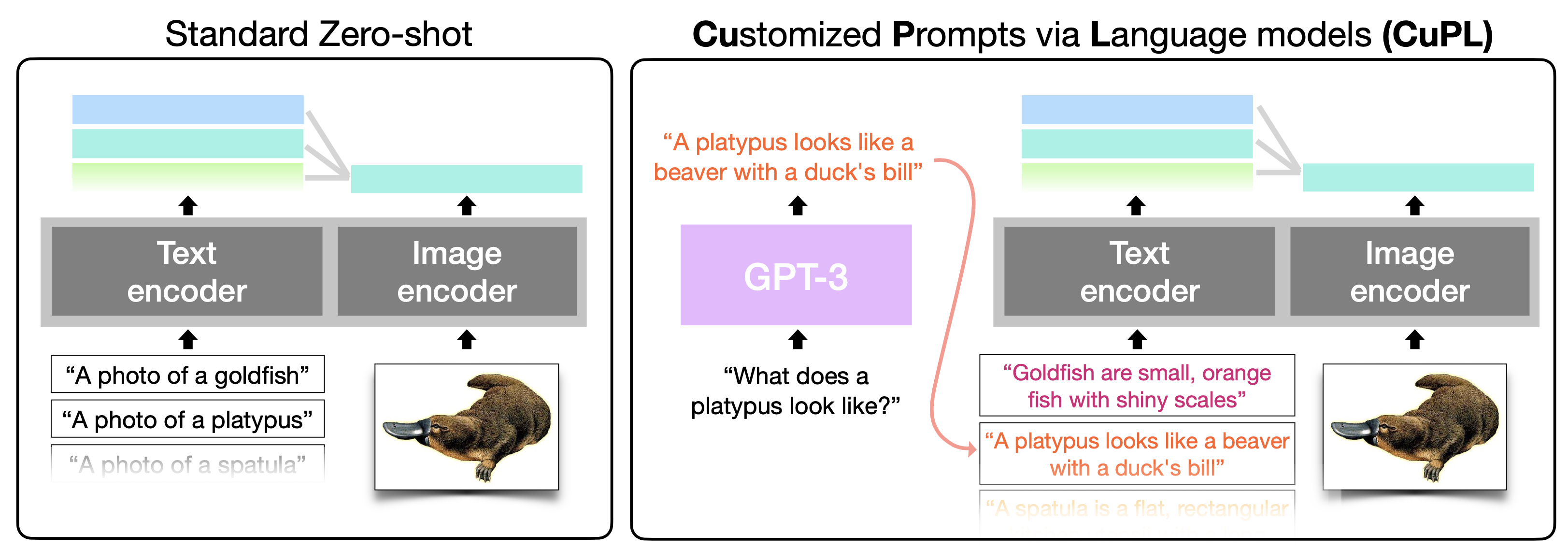}
        \caption[The CuPL Architecture]{The CuPL Architecture \citep{pratt2023}}
    \end{center}
\end{figure}

\textbf{\textit{Multimodal Prompting with Missing Modalities for Visual Recognition}} \citep{lee2023multimodal} attempts to reduce the effect of missing modalities on classification accuracy, an issue that can introduce bias into multimodal transformers. Despite this, a significant limitation was that they could not recover missing information from the multimodal inputs. We intend to solve this issue by synthesizing data to better represent the true distribution of the source, improving the accuracy of the model’s prediction \citep{park2018data}.

\begin{figure}[h]
    \begin{center}
        \includegraphics[width=14cm]{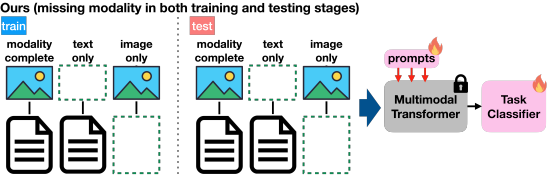}
        \caption[Framework to Handle Missing Modalities]{Framework to Handle Missing Modalities \citep{lee2023multimodal}}
    \end{center}
\end{figure}

\begin{figure}[H]
    \begin{center}
        \includegraphics[width=9cm]{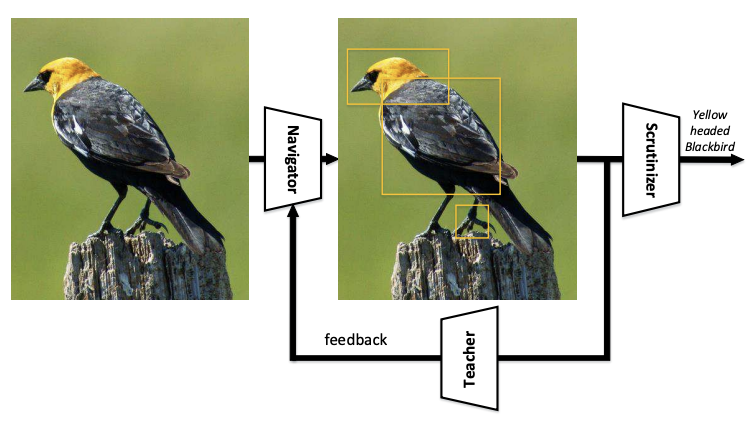}
        \caption[The NTSNet Architecture]{The NTSNet Architecture \citep{yang2018learning}}
        \label{fig:ntsnet}
    \end{center}
\end{figure}

A model that applies this concept of model ensembling was introduced in \textbf{\textit{Learning to Navigate for Fine-grained Classification}} by \citet{yang2018learning}. This is a state-of-the-art paper that attempts to reduce misclassification rates by developing a model called NTS-Net (Navigator-Teacher-Scrutinizer Network) to teach itself methods of identifying and scrutinizing fine-grained image details. The Navigator navigates the model to focus on the most informative regions (denoted by yellow rectangles in \cref{fig:ntsnet}), while the Teacher evaluates the regions proposed by the Navigator and provides feedback. After that, the Scrutinizer scrutinizes those regions to make predictions. NTS-Net achieves high classification accuracy on a pre-defined dataset; however, we wanted to explore if these foundational concepts could be applied to a training-free, zero-shot environment. In order to achieve this, we leverage pretrained open-vocabulary models with advanced attention mechanisms to discriminate the fine-grained features of a given image, in order high performance on a variety of classes without additional training.

\section{Methods}
In this section, we explore and analyze our selection of datasets, models, the MuSE Architecture, and implications of the final results.

\subsection{Datasets}
Within this chapter, we specifically select four datasets that will test the generalizability and fine-grained classification capabilities of MuSE: Flowers 102 \citep{Nilsback08}, Describable Textures Dataset (DTD) \citep{cimpoi14describing}, Remote Sensing Image Scene Classification 45 (RESISC45) \citep{cheng2017remote}, and Fine-Grained Visual Classification (FGVC) Aircraft \citep{maji13fine-grained}. We selected these datasets because, alongside testing generalizability and fine-grained classification, they were extremely diverse. Each dataset contains completely different information from the others, which will allow us to understand how robust this technique is to distribution shift.\\
\\
Flowers 102 contains images split into 102 categories, with large scale, pose, and light variations between each image. There are at least 40 images per category, which were collected by searching the internet and manually taking pictures. We focus on the category selection because it is particularly useful for our evaluation. Within the dataset, some categories have large variations, but also others that are very similar to one another. This is useful because many factors contribute to distinguishing one flower from another. This means we can evaluate the model on general and fine-grained classification capabilities.\\
\\
DTD contains 47 classes, 5,650 images total, and has similar benefits as Flowers 102; however, it additionally allows us to evaluate the generalizability of our technique. Since many of the category titles are quite abstract, the model must be able to generalize the class name to any images conforming to the proper description. This is especially difficult since certain classes are similar to one another, so the model must have a deep cross-modal representation of these classes.\\
\\
RESISC45 contains 45 scene classes, with 700 images per class and 31,500 images total. The dataset contains images of satellite photos containing different scenes, and each category holds significant variation in translation, spatial resolution, viewpoint, object pose, illumination, background, and occlusion. Scene classification is a particularly difficult task because generating images for scenes is less precise than generating specific subjects. To analyze the limitations and drawbacks of MuSE, we must explore if the technique can still improve performance on this dataset.\\
\\
Finally, FGVC Aircraft contains 100 aircraft models organized in a three-level hierarchy, with 10,000 total images. The hierarchy of aircraft within this dataset is key to our evaluation. The other datasets focus on a balance between fine and coarse-grained classification; however, most FGVC Aircraft are fine-grained. The hierarchical structure means we can better explore the per-class accuracies and evaluate which granularity the technique breaks down.

\subsection{Models}
Within CuPL, GPT-3 DaVinci-002 was used to generate the prompts, CLIP ViT-L/14 is used as MuSE's multimodal model for image-to-text retrieval, and Stable Diffusion XL 1.0 is utilized for image generation.

\subsection{The MuSE Architecture}
\begin{figure}[H]
    \begin{center}
        \includegraphics[width=16.5cm]{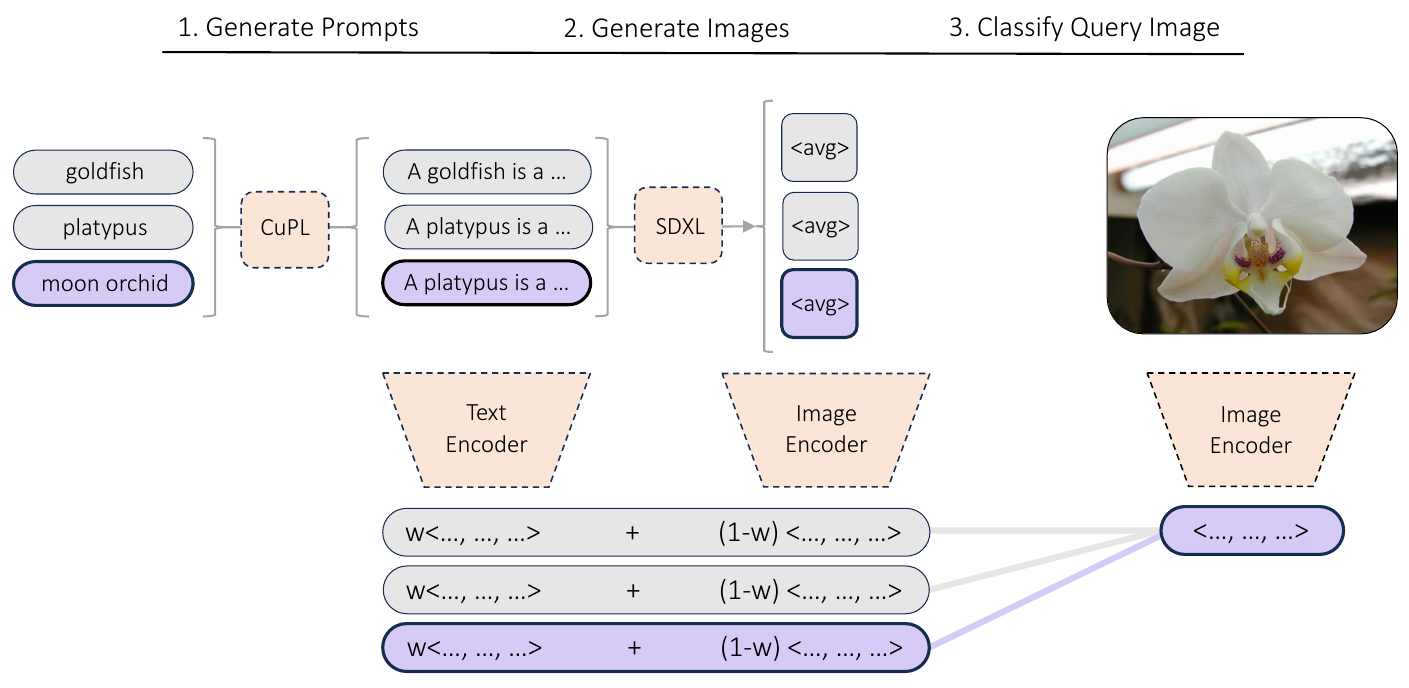}
        \caption[The MuSE Architecture]{The MuSE Architecture}
        \label{fig:muse}
    \end{center}
\end{figure}
\sloppy
The MuSE architecture is as follows:\\
\\
\textbf{Step 1: Generate Prompts}\\
The first stage is to leverage CuPL to create contextually relevant prompts for our multimodal model. CuPL is designed to generate prompts that effectively guide the model's attention and understanding, enabling better performance in zero-shot image classification. It does so by prompting a Large Language Model (LLM) to generate a description of a given class. These descriptions emphasize the visual information of each class so that when the multimodal model classifies the image, it can utilize the richer text to get a much more accurate embedding. This is imperative because using an LLM allows us to quickly generate many descriptive prompts customized to each category. In the MuSE architecture \cref{fig:muse}, this corresponds to the step where each class name is passed into the CuPL/SDXL block. For MuSE, we simply generate one prompt per class and then use that prompt in the next step.
\\
\\
\textbf{Step 2: Generate Images}\\
We now leverage the prompts generated by CuPL in the previous stage. We simply use each prompt to generate an image for each class. As previously stated, Stable Diffusion XL (SDXL) 1.0 is used in order to generate our images. For every image, we use 50 generation steps, a guidance scale of 15, and a seed of 0. We generate 5 images per class using the same prompt, then average the normalized embeddings of these images as shown in \cref{fig:muse}.
\\
\\
\textbf{Step 3: Classify Query Image}\\
Using both the generated prompts and generated images, we classify the query image. For each class, we take the embedding from the generated prompt and the embedding from the generated image, then create a weighted sum of their normalized embeddings. We determine the weight for this sum, by scanning values from 0 to 1 by increments of 0.01, where the text is multiplied by weight $w$ and the images are multiplied by weight $(1 - w)$. Upon creating a joint embedding, we then use cosine similarity to classify the query image.

\section{Results}
\label{sec:muse_results}
\subsection{Metrics}
We will focus on accuracy for our evaluations; however, the specific type of accuracy used will vary based on the dataset's standards. The metrics are as follows:\\
\begin{table}[h]
    \begin{center}
        \begin{tabular}{|c|c|c|c|}
            \hline
             \textbf{Flowers 102} & \textbf{DTD} & \textbf{RESISC45} & \textbf{FGVC Aircraft}\\
            \hline
            Mean Per-Class Accuracy & Top-1 Accuracy & Top-1 Accuracy & Mean Per-Class Accuracy\\
            \hline
        \end{tabular}
        \caption{Evaluation Metrics for Each Dataset}
        \label{tab:eval_metrics}
    \end{center}
\end{table}\\
As alluded to within the analysis of our datasets, the evaluation of our approach is multifaceted. It focuses on both traditional image classification benchmarks and fine-grained image classification. The multiple accuracy metrics will also allow us to understand if the method improves performance across all classes or boosts the top 1.

\subsection{Evaluation Breakdown}
\begin{figure}[h]
    \begin{center}
        \includegraphics[width=11cm]{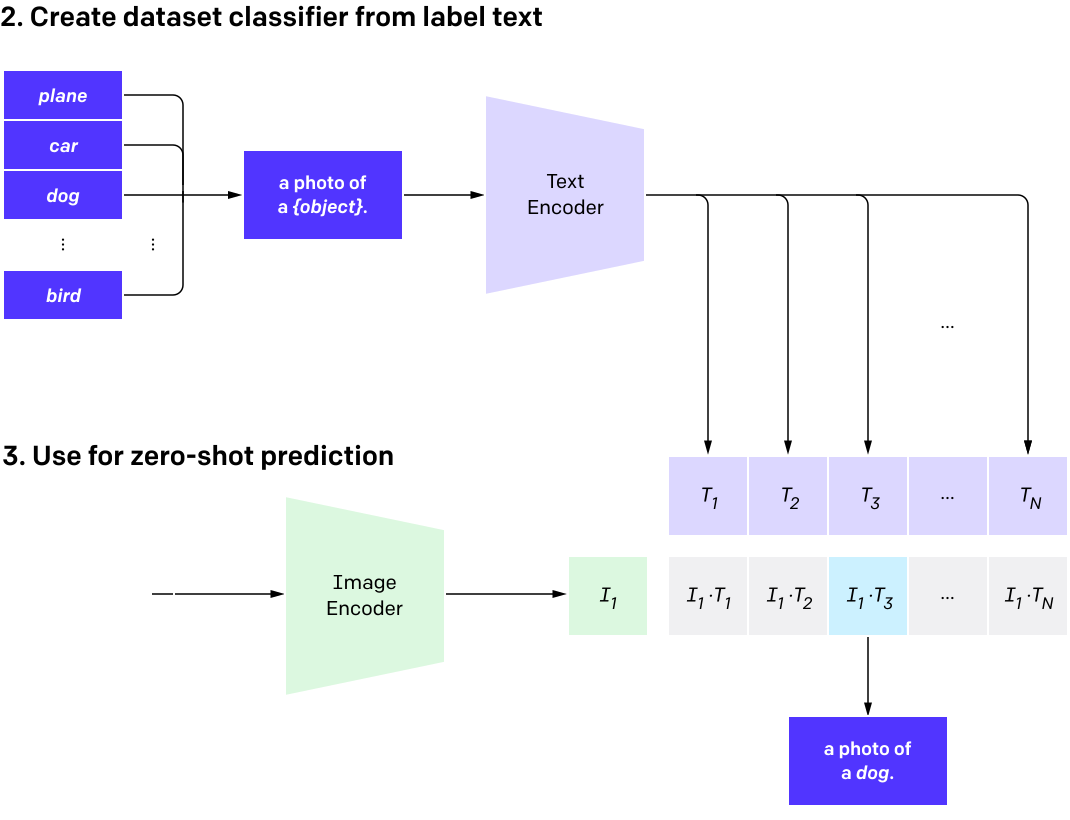}
        \caption[Classifying an image with CLIP at inference-time]{Classifying an image with CLIP at inference-time \citep{radford2021learning}.}
        \label{fig:muse_clip}
    \end{center}
\end{figure}

\subsubsection{Baselines}
The baseline will be the standard method of CLIP image classification as in \cref{fig:muse_clip}, where the image is encoded into embedding space and then matched with each of the class embeddings to reveal the highest score.\\
\\
We use a variety of prompts and templates for this CLIP baseline. Note that within the text being used to get cosine similarities for the CLIP baseline, is the same text that is immediately after the CuPL block in \cref{fig:muse}. This text can either be the prompt template "A photo of a <classname>" or a single CuPL prompt per class.\\
\\
In addition to our standard CLIP baseline, we also provide the classification accuracy when only generated images are used instead of text. This is equivalent to $w = 0$ and thus $(1 - w) = 1$. We provide this metric to show why images alone are not sufficient for improved classification, and that multiple modalities are needed for effective classification.

\subsubsection{MuSE Evaluations}
For MuSE, we construct a variety of different evaluations. Our first is the standard method of classification shown in \cref{fig:muse}. We use CuPL to generate a prompt for each class, then generate 5 images based on that prompt. We average the normalized embeddings of all five images, then create a weighted sum of the embeddings from the desired text and generated images. The text being used can vary depending on our evaluation; however, it is important to note that this method is characterized by its weighting strategy. \\
\\
We define matrices $t = [t_1, t_2, ..., t_N]$ and $i = [i_1, i_2, ..., i_N]$ as containing the normalized embeddings for the generated text and images, respectively. Similarly, matrix $q$ is defined as the vector containing the normalized query image embedding for a single query image. Both $t$ and $i$ are of dimensions $N \text{x} M$, where $N$ is the total number of classes within the dataset, and $M$ is the length of a single embedding vector. For the CLIP ViT-L/14 text and image encoders, $M = 768$. Similarly, $q$ is of shape $1 x M$, because we embed one query image at a time. Possible weight values are scanned from 0 to 1 by steps of 0.01, where the text is weighted by weight $w$, and the images are weighted by $(1 - w)$.\\
\\
Upon scanning all values, we take the maximum top-1 accuracy and use that weight configuration for our final result. The full formulation to get our final combined embedding $x_{\text{Std}}$ using our Standard MuSE method, and given a single query image $q$ is defined as:\\
$$x_{\text{Std}} = \underset{f(w)}{\mathrm{argmax}\text{ }s(f(w))} = \{ f(w) | (s(f(w)) < s(f(w_j))) \forall \{w_j | w_j \in \mathbb{R}, 0 \leq w_j \leq 1\}\}$$\\
where\\
$$f(w) = (w \cdot t + (1 - w) \cdot i)$$ and $$s(f(w)) = f(w) \cdot q$$\\
\\
Since we normalize all our embeddings, we can simply use the dot product to retrieve our cosine similarity.\\
\\
Our second MuSE evaluation method utilizes confidence scores and previously described standard weight values. Where $w \in [0, 1]$, the confidence score $c_i$ represents how likely the image belongs to the class at index $i$, is the true class of the query image $q$. We define\\
$$c_i = 1 - \frac{e^{q \cdot t_i}}{\sum_{j=1}^N e^{q \cdot t_j}}$$\\
Notice how we use the inverse confidence score because we calculate the score using text embeddings. We want our resulting confidence to represent how likely the image belongs to the class at index $i$, so we take the inverse text confidence so we don't rely too heavily on text or images. If we are not confident our text is correct, we will utilize more of the image embeddings, and vice versa. We leverage $c_i$ to create our new formulation for our final combined embedding $x_{\text{Conf}}$ utilizing this Confidence MuSE method:\\
$$x_{\text{Conf}} = \underset{f(w)}{\mathrm{argmax}\text{ }s(f(w))} = \{ f(w) | (s(f(w)) < s(f(w_j))) \forall \{w_j | w_j \in \mathbb{R}, 0 \leq w_j \leq 1\}\}$$\\
where\\
$$f(w) = (w \cdot t + (1 - w) \cdot c_t \cdot i)$$ and $$s(f(w)) = f(w) \cdot q$$\\
\\
Our third MuSE evaluation tests the performance while setting a fixed weight value. This is the exact same formulation as MuSE Standard, but $w = 0.1$ so we do not scan for the argmax. Similarly, our final MuSE evaluation sets $w = 0.1$ for MuSE Confidence.

\subsection{Classification Results}
We start by evaluating the results with the text being "A photo of a <classname>." We still use the CuPL text in order to generate our images; however, we replace the text embeddings with the normalized "A photo of a <classname>" embedding as shown in the bottom of \cref{fig:muse}.

\begin{table}[h]
    \begin{center}
        \begin{tabular}{|c|c|c|c|c|}
            \hline
             & \textbf{Flowers 102} & \textbf{DTD} & \textbf{FGVC Aircraft} & \textbf{RESISC45}\\
            \hline
            \rowcolor{gray!20} 
            \cellcolor{white!10}CLIP & 75.53 & 51.54 & 31.22 & 61.43\\
            \hline
            \rowcolor{red!10} 
            \cellcolor{white!10}Images Only & 48.21 & 29.02 & 15.64 & 36.72\\
            \hline
            MuSE Standard & 76.83 & \cellcolor{green!10}52.02 & 31.97 & 63.69\\
            \hline
            MuSE Confidence & \cellcolor{green!10}76.85 & 51.99 & \cellcolor{green!10}32.06 & \cellcolor{green!10}63.73\\
            \hline
            MuSE Standard $w = 0.1$ & 76.55 & 51.99 & 31.97 & 63.35\\
            \hline
            MuSE Confidence $w = 0.1$ & 76.57 & 51.99 & \cellcolor{green!10}32.06 & 63.34\\
            \hline
        \end{tabular}
        \caption{MuSE's accuracy using "A photo of a <classname>" as the text embedding.}
        \label{tab:MuSE_a-photo_results}
    \end{center}
\end{table}

From these results, we can see that MuSE already performs better than the baselines across \textbf{all} of our datasets. The improvements range from 1-2\% increases, but this is still impressive considering the variety of the datasets used. We gain a deeper understanding as we examine other evaluations.

\begin{table}[h]
    \begin{center}
        \begin{tabular}{|c|c|c|c|c|}
            \hline
             & \textbf{Flowers 102} & \textbf{DTD} & \textbf{FGVC Aircraft} & \textbf{RESISC45}\\
            \hline
            \rowcolor{gray!20} 
            \cellcolor{white!10} CLIP & 78.59 & 55.18 & 32.85 & 69.86\\
            \hline
            \rowcolor{red!10} 
            \cellcolor{white!10} Images Only & 48.21 & 29.02 & 15.64 & 36.72\\
            \hline
            MuSE Standard & \cellcolor{green!10}79.26 & \cellcolor{green!10}55.46 & \cellcolor{green!10}33.68 & \cellcolor{green!10}70.23\\
            \hline
            MuSE Confidence &\cellcolor{green!10} 79.26 & 55.44 & 33.65 & \cellcolor{green!10}70.23\\
            \hline
            MuSE Standard $w = 0.1$ & 79.18 & \cellcolor{green!10}55.46 & \cellcolor{green!10}33.68 & 70.07\\
            \hline
            MuSE Confidence $w = 0.1$ & 79.20 & 55.44 & 33.65 & 70.10\\
            \hline
        \end{tabular}
        \caption{MuSE's accuracy using an average of CLIP templates as the text embedding.}
        \label{tab:MuSE_avg-clip_results}
    \end{center}
\end{table}

For this evaluation, we started by obtaining the CLIP evaluation templates, originally published in the paper by \cite{radford2021learning}. We provide the list of all CLIP templates used within \cref{appendix:muse_appendix} \cref{tab:clip_templates}. For every class in the dataset, we average the embeddings of each of the CLIP prompts that the dataset contains (e.g., 'a photo of a \{\}, a type of flower.'). We then use the averaged embeddings in the exact same way that we use the standard text embeddings, by creating a weighted sum and then classifying as shown in \cref{fig:muse}. The results of this technique are shown above in \cref{tab:MuSE_avg-clip_results}. We can see that once again, MuSE improves performance across all datasets; however, another important feature of this evaluation, is the fact that two of the evaluations with MuSE Standard where $w = 0.1$, ended up being the optimal configuration. This suggests that we might be able to set a fixed value based on the specific evaluation, and still achieve optimal performance. We further explore this idea in the next evaluation.

\begin{table}[h]
    \begin{center}
        \begin{tabular}{|c|c|c|c|c|}
            \hline
             & \textbf{Flowers 102} & \textbf{DTD} & \textbf{FGVC Aircraft} & \textbf{RESISC45}\\
            \hline
            \rowcolor{gray!20} 
            \cellcolor{white!10}CLIP & 73.32 & 51.54 & 31.59 & 56.26\\
            \hline
            \rowcolor{red!10} 
            \cellcolor{white!10}Images Only & 48.21 & 29.02 & 15.64 & 36.72\\
            \hline
            MuSE Standard & 77.08 & \cellcolor{green!10}52.02 & \cellcolor{green!10}32.55 & 58.34\\
            \hline
            MuSE Confidence & \cellcolor{green!10}77.12 & 51.99 & 32.47 & \cellcolor{green!10}58.39\\
            \hline
            MuSE Standard $w = 0.1$ & 75.54 & 51.99 & \cellcolor{green!10}32.55 & 57.56\\
            \hline
            MuSE Confidence $w = 0.1$ & 75.50 & 51.99 & 32.47 & 57.58\\
            \hline
        \end{tabular}
        \caption{MuSE's accuracy using the CuPL prompt as the text embedding.}
        \label{tab:MuSE_cupl_results}
    \end{center}
\end{table}

Within this evaluation, we have the exact same setup as the "A photo of a <classname>" text prompt; however, we replace that text with a single CuPL prompt. Since CuPL prompts are much more visually descriptive, we expect the new text embedding to better represent the true class. Upon analyzing the results in \cref{tab:MuSE_cupl_results}, we can see that our hypothesis was correct. Within this evaluation, we observe accuracy gains up to 3\%. This is extremely significant given the zero-shot, training-free, and minimal overhead nature of our approach. Even when viewing the sub-optimal MuSE results, we can see that they still are able to achieve significantly higher accuracies when compared to the baselines. Despite this, when viewing the next evaluation, we are also able to see when this method fails.

\begin{table}[h]
    \begin{center}
        \begin{tabular}{|c|c|c|c|c|}
            \hline
             & \textbf{Flowers 102} & \textbf{DTD} & \textbf{FGVC Aircraft} & \textbf{RESISC45}\\
            \hline
            \rowcolor{gray!20} 
            \cellcolor{white!10}CLIP & 79.27 & 55.18 & 35.75 & 71.45\\
            \hline
            \rowcolor{red!10} 
            \cellcolor{white!10}Images Only & 48.21 & 29.02 & 15.64 & 36.72\\
            \hline
            MuSE Standard & 80.11 & \cellcolor{green!10}55.46 & 36.17 & \cellcolor{gray!20}71.45\\
            \hline
            MuSE Confidence & \cellcolor{green!10}80.13 & 55.44 & \cellcolor{green!10}36.19 & \cellcolor{gray!20}71.45\\
            \hline
            MuSE Standard $w = 0.1$ & 80.11 & \cellcolor{green!10}55.46 & 36.17 & \cellcolor{red!10}71.06\\
            \hline
            MuSE Confidence $w = 0.1$ & \cellcolor{green!10}80.13 & 55.44 & 36.08 & \cellcolor{red!10}71.08\\
            \hline
        \end{tabular}
        \caption{MuSE's accuracy using an average of many CuPL prompts as the text embedding.}
        \label{tab:MuSE_avg-cupl_results}
    \end{center}
\end{table}

\begin{figure}[h]
    \begin{center}
        \includegraphics[width=8cm]{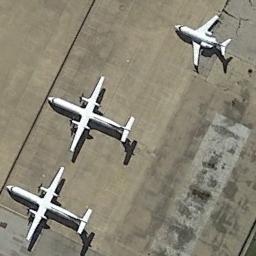}
        \caption{A sample RESISC45 ground-truth airplane.}
        \label{fig:resisc45_airplane}
    \end{center}
\end{figure}

In this evaluation in \cref{tab:MuSE_avg-cupl_results}, we utilize anywhere from 20-60 CuPL prompts per class, depending on the dataset. We can still see improvements; however, we also see the first negative results on RESISC45. Within this dataset, we are not able to get any performance increase, and there even is a decrease in accuracy when we set constant weight values. The reason why this occurs, is due to the quality of the CuPL prompts used. We can see a sample of the CuPL prompts used for this evaluation, within \cref{tab:sample_cupl_prompts}. When viewing the prompts from RESISC45, we can see they are still of good quality; however, we notice issues when comparing the prompt to ground truth images within the dataset. The prompts typically describe airplanes mid-flight, here are a few of the prompts for example:
\begin{itemize}
    \item "A satellite photo of an airplane might show the plane in flight, with its wings outstretched and its engines propelling it through the air."
    \item "The photo might show the airplane in mid-flight, with the sun's rays shining off its metal body."
    \item "A satellite photo of an airplane would show a plane in flight, with its wings outstretched and its body elongated."
\end{itemize}
These descriptions, while not incorrect, are nothing like the ground truth within the dataset. We can clearly see this when viewing \cref{fig:resisc45_airplane}\\
\\
The images containing airplanes within the dataset, are all airplanes located on the ground. This is a key issue and one we discuss later in this chapter. The fact that the images generated can be completely accurate to the prompt, and yet the ground truth image could be different, is one of the major difficulties with this technique. To further understand how these issues occur, view the sample generated images compared to their texts within \cref{appendix:d3g_appendix}.

\subsection{Fine-Grained Classification}
\begin{figure}[H]
    \begin{center}
        \includegraphics[width=12cm]{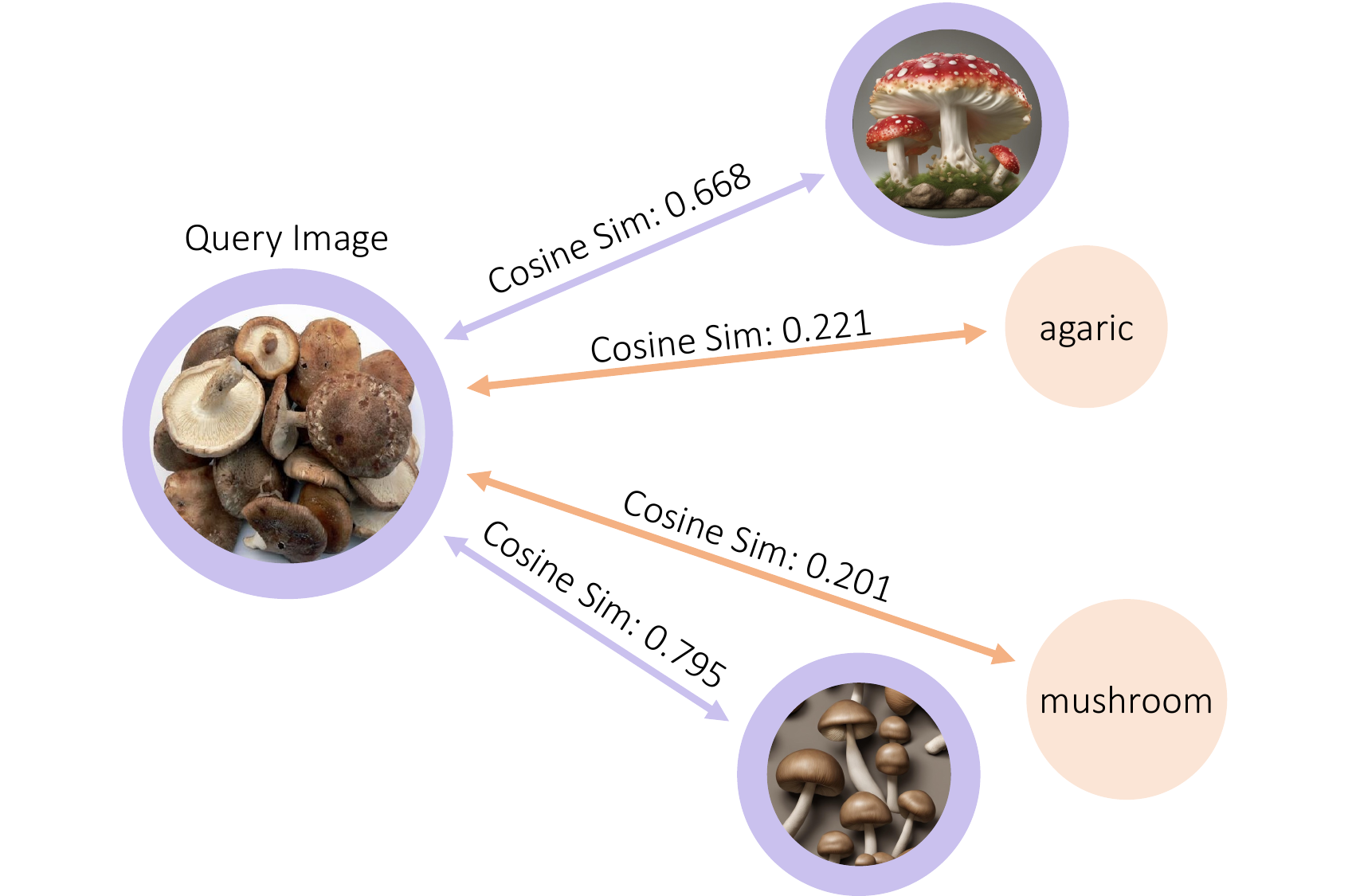}
        \caption{How MuSE Improves Fine-Grained Classification}
        \label{fig:muse_fine_grained_diagram}
    \end{center}
\end{figure}
To further explore MuSE's impact on fine-grained classification accuracies, we tested the binary classification accuracy on similar classes within ImageNet. We selected this dataset because we wanted to identify a large commonly used dataset, then isolate the mistakes CLIP was making on fine-grained classes. To identify these fine-grained classes, we selected the true classes that CLIP predicted incorrectly most often and took the corresponding CLIP predictions for these classes. This provided a subset that exposed CLIP's inaccuracies on ImageNet.\\
\\
Within this dataset, mushroom and agaric were a commonly confused pair of classes, with the misclassification rate for mushroom being extremely high. To understand why this was the case, we compared the CLIP and MuSE performance, as shown in \cref{tab:fine_grained_acc}.
\begin{table}[H]
    \begin{center}
        \begin{tabular}{|c|c|c|}
            \hline
             & \textbf{Mushroom} & \textbf{Agaric}\\
            \hline
            CLIP & 0.1 & 1.0\\
            \hline
            \rowcolor{green!20} MuSE (average) & 0.56 & 0.9\\
            \hline
        \end{tabular}
        \caption{Fine Grained Accuracy over the Mushroom and Agaric classes}
        \label{tab:fine_grained_acc}
    \end{center}
\end{table}
CLIP is overfitting, because it cannot distinguish between the two categories. This is why the model achieves perfect accuracy on agaric while rarely selects mushroom. In contrast, MuSE balances these numbers by dramatically raising the accuracy on the mushroom class while still achieving extremely high accuracy on the agaric class, showing it can distinguish between the classes and still often select the correct class. We demonstrate this within \cref{fig:muse_fine_grained_diagram}. In this figure, CLIP initially misclassifies the query image with a ground-truth label of "mushroom" to be an "agaric" as shown by the orange arrows. The higher similarity score combined with the overfitting demonstrated in \cref{tab:fine_grained_acc} sheds light on the issue. Despite this, we can clearly see that by generating images of each class, then creating a weighted sum of those new embeddings, we can achieve a correct classification as shown by the purple arrows in the diagram.\\
\\
We ran tests on a many other classes, and achieved similar results. The accuracy improvement for the categories "Frilled Lizard" and "Agama" are shown below in \cref{tab:fine_grained_acc2}
\begin{table}[H]
  \centering
  \begin{tabular}{|c|c|c|}
    \hline
     & \textbf{Frilled Lizard} & \textbf{Agama}\\
    \hline
    CLIP & 1.0 & 0.78\\
    \hline
    \rowcolor{green!20} MuSE (offset=0.1) & 0.94 & 0.84\\
    \hline
  \end{tabular}
  \caption{Fine Grained Accuracy over the Frilled Lizard and Agama classes}
  \label{tab:fine_grained_acc2}
\end{table}

\subsection{Class Imbalances}
Finally, we would like to bring attention to a final class, the Bishop of Llandaff. This is a class from the Flowers 102 dataset that showcased a final benefit of MuSE. This is because our CLIP baseline received 0\% accuracy on classifying this particular flower. That means CLIP had very little cross modal representations of the Bishop of Llandaff, resulting in very little understanding of the flower. This broader issue arises from class imbalance within the training data. As mentioned within the Introduction (\cref{chapter:intro}), multimodal models \textbf{must} be trained on datasets with rich cross-modal representations. The reason why, is because without these rich representations, class imbalances will occur. The issue is that no dataset can realistically have rich cross-modal representations for every possible category, so this is where MuSE comes in. Even if the generative model, or even the classifier model does not fully understand a given class, we can leverage CuPL to generate a description that breaks the concept down into its simpler parts, then generate an image that composes each of these parts. This is better demonstrated with an example:\\
\begin{figure}[H]
    \begin{center}
        \includegraphics[height=6cm]{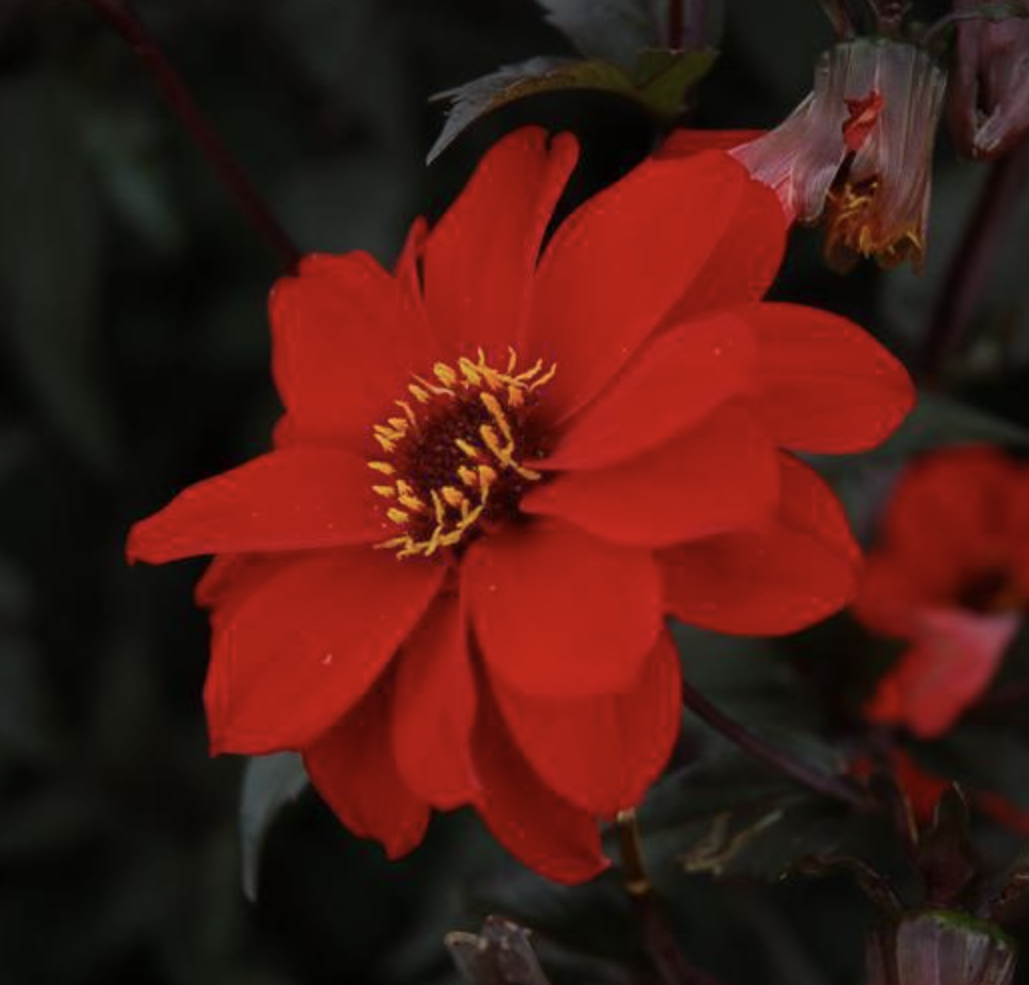}
        \includegraphics[height=6cm]{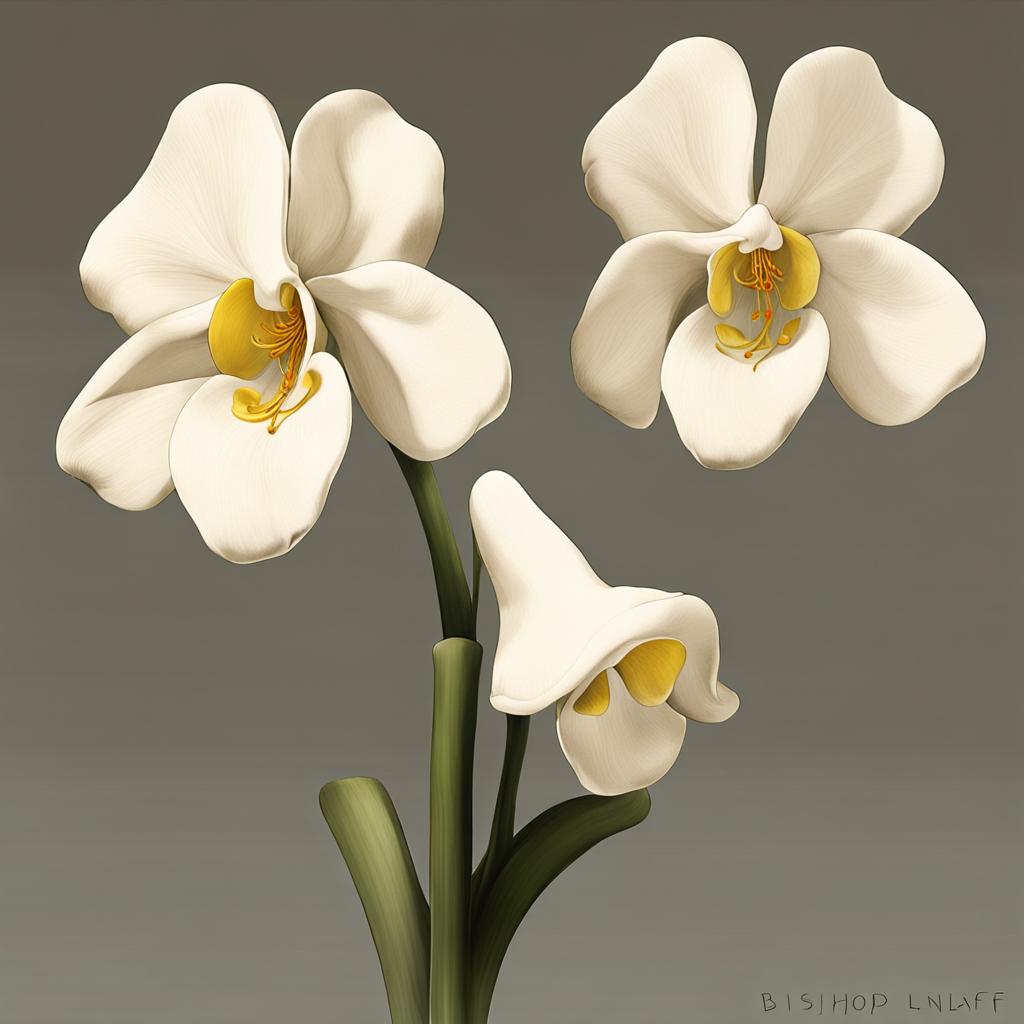}
        \caption{Ground Truth Bishop of Llandaff (left) compared to a generated image of the Bishop of Llandaff (right)}
        \label{fig:muse_bishop}
    \end{center}
\end{figure}
In \cref{fig:muse_bishop} we see a ground truth photo of the Bishop of Llandaff, compared to a Stable Diffusion XL generated photo of the same class. This clearly shows that the model does not understand what the class is. In this scenario, MuSE will not improve the baseline CLIP performance at all; however, this can change with a few small adjustments. If we customize our CuPL prompt to include a series of intermediate steps (similar to chain-of-thought), we can form a new image based predominantly on the description provided and not the model's knowledge of the class. This is demonstrated by the new image generated with this technique:\\
\begin{figure}[H]
    \begin{center}
        \includegraphics[height=6cm]{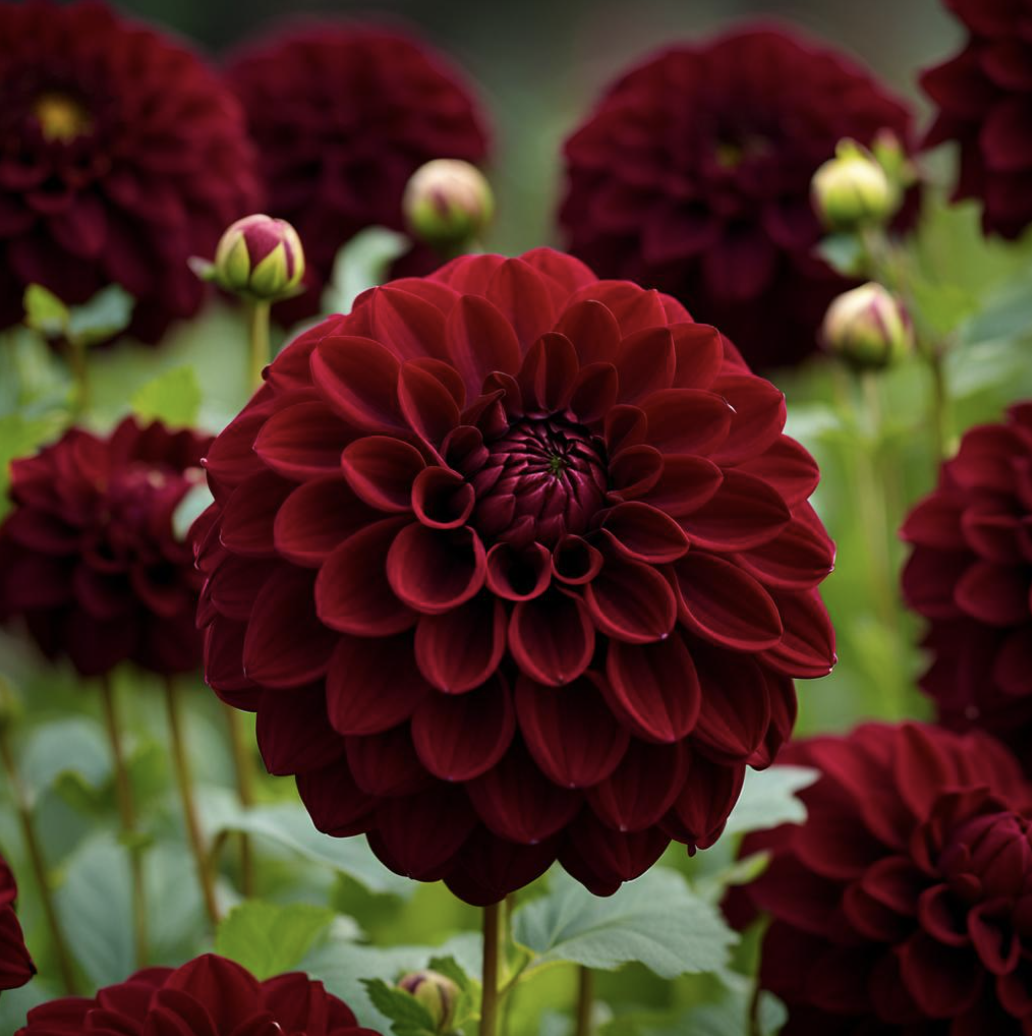}
        \caption{Improved generated image of the Bishop of Llandaff}
        \label{fig:muse_good_gen_bishop}
    \end{center}
\end{figure}
For this image, our CuPL prompt was "Dahlia 'Bishop of Llandaff' is a cultivar known for its dark red or maroon flowers and dark foliage." Even though it is similar in structure to the other CuPL prompts, it contains a few key details:

\begin{itemize}
    \item It includes the hierarchical class structure within the prompt. The Bishop of Llandaff is a type of Dahlia, so including this extra information allows the model to infer what the flower should look like.
    \item The prompt also describes additional visual features which SDXL uses to improve its image generation. Features such as color or foliage descriptions all help the generation process.
\end{itemize}

The fact that modifying the prompt allowed the model to generate a class it did not understand, is an idea that we will further explore in \cref{sec:future_work}. This simple adjustment allowed use to get the resulting accuracies on the Bishop of Llandaff.

\begin{table}[H]
  \centering
  \begin{tabular}{|c|c|}
    \hline
     & \textbf{Bishop of LLandaff}\\
    \hline
    CLIP & 0 \\
    \hline
    \rowcolor{green!20} Standard MuSE & 0.45 \\
    \hline
  \end{tabular}
  \caption{Accuracy on the Bishop of Llandaff}
  \label{tab:fine_grained_acc3}
\end{table}

\section{Discussion}
As is evident in all of the results, but most prominently within \cref{tab:MuSE_a-photo_results}, using solely images does not work at all for this method of image classification. A major reason for this is due to the fact that the performance of MuSE is heavily dependent on whether the generated images somewhat match the distribution of images in the dataset (i.e. whether or not the image generation model can generate a good image of a given class). Similarly, it's possible that CLIP does not understand the difference between fine-grained classes (or at least it doesn't understand these differences very well). This can easily occur due to a non-comprehensive training set, or simply rare classes being used for evaluation. \\
\\
With this in mind there are two prominent assumptions made in this proposal: 
\begin{itemize}
    \item It is assumed that generative models used will better represent the true distribution of the data (due to their increased complexity and data diversity), and the multimodal model can distinguish between similar classes. If the multimodal model has learned separate classes are semantically identical, MuSE will not improve performance because there is a flaw in the fundamental knowledge of the multimodal model. If successful, this research would leverage the capabilities of expert models without retraining to gain data diversity, making multimodal systems more robust and capable of handling real-world challenges.
    
    \item It is important to note that despite aiming to improve multimodal fairness, MuSE \textbf{does not} produce fair results. Instead, it offsets learned biases. This means that MuSE can either reduce or accentuate human bias, and should not be used as a universal architecture to improve multimodal model fairness. This will be further explored in later chapters.
\end{itemize}

\section{Conclusion}
Traditional multimodal classification methods face issues when dealing with fine-grained classes due to limited data representation. These challenges can lead to underfitting and misclassification. To mitigate these issues we have proposed MuSE, which utilizes generative prompting to improve image classification accuracy across a range of datasets, making it a highly generalizable technique. This approach is zero-shot, training-free, contains minimal overhead, and is able to reduce the misclassification rate while enhancing the accuracy of any multimodal classification model (if it confines to the restrictions mentioned within the Discussion section). The results shown highlight the significant impact MuSE can have on image classification, and on many future downstream tasks.

\chapter{D3G}
\label{chapter:d3g}
\section{Overview}
Image classification is a task essential for machine perception to achieve human-level image understanding. Multimodal models such as CLIP, have been able to perform well on this task by learning semantic similarities across Vision and Language; however, despite these  advances, image classification is still challenging task. Models with low capacity often suffer from underfitting and thus underperform on fine-grained image classification. Along with this, it is important to ensure high-quality data with rich cross-modal representations of each class, which is often difficult to generate. When datasets do not enforce balanced demographics, the predictions will bias toward the more represented class, while others will be neglected. We focus on how these issues can lead to harmful bias for zero-shot image classification, and explore how to combat these issues in demographic bias. We propose Diverse Demographic Data Generation (D3G), a training-free, zero-shot method of boosting classification accuracy while reducing demographic bias in pre-trained multimodal models. With this method, we utilize CLIP as our base multimodal model, and Stable Diffusion XL as our generative model. We demonstrate providing diverse demographic data at inference time improves performance for these models, and explore the impact of individual demographics on the resulting accuracy metric.

\section{Introduction}
Deep Learning systems have been a promising new paradigm in the field of image classification. Vision-Language systems are able to utilize multiple modalities to create models that are more generalizable on a broad range of downstream tasks. Despite this performance, issues such as data redundancy, noise, and class imbalance are just a few of the many difficulties that arise from collecting large amounts of training data. Multimodal models in particular, require large amounts of high-quality training data with rich cross-modal representations in order to perform well compared to their unimodal counterparts. These models leverage the massive amounts of image text pairs available online by learning to associate the images with their correct caption, leading to greater flexibility during inference \cite{pratt2023}; however, class imbalances can often lead to gender and racial bias depending on the desired task.\\
\\
Existing public face image datasets are strongly biased toward Caucasian faces, meanwhile other races (i.e., Latino) are significantly underrepresented \cref{fig:face_datasets}. As a result, the models training from such datasets suffer from inconsistent classification accuracies, limiting the applicability of such systems work predominantly on White racial groups \citet{karkkainen2021fairface}. This means that minority sub-populations can potentially be further marginalized when applied to certain downstream tasks without calibration. This is a core tenet of machine learning: biased data produces biased models.
\begin{figure}[H]
    \begin{center}
        \includegraphics[width=7cm]{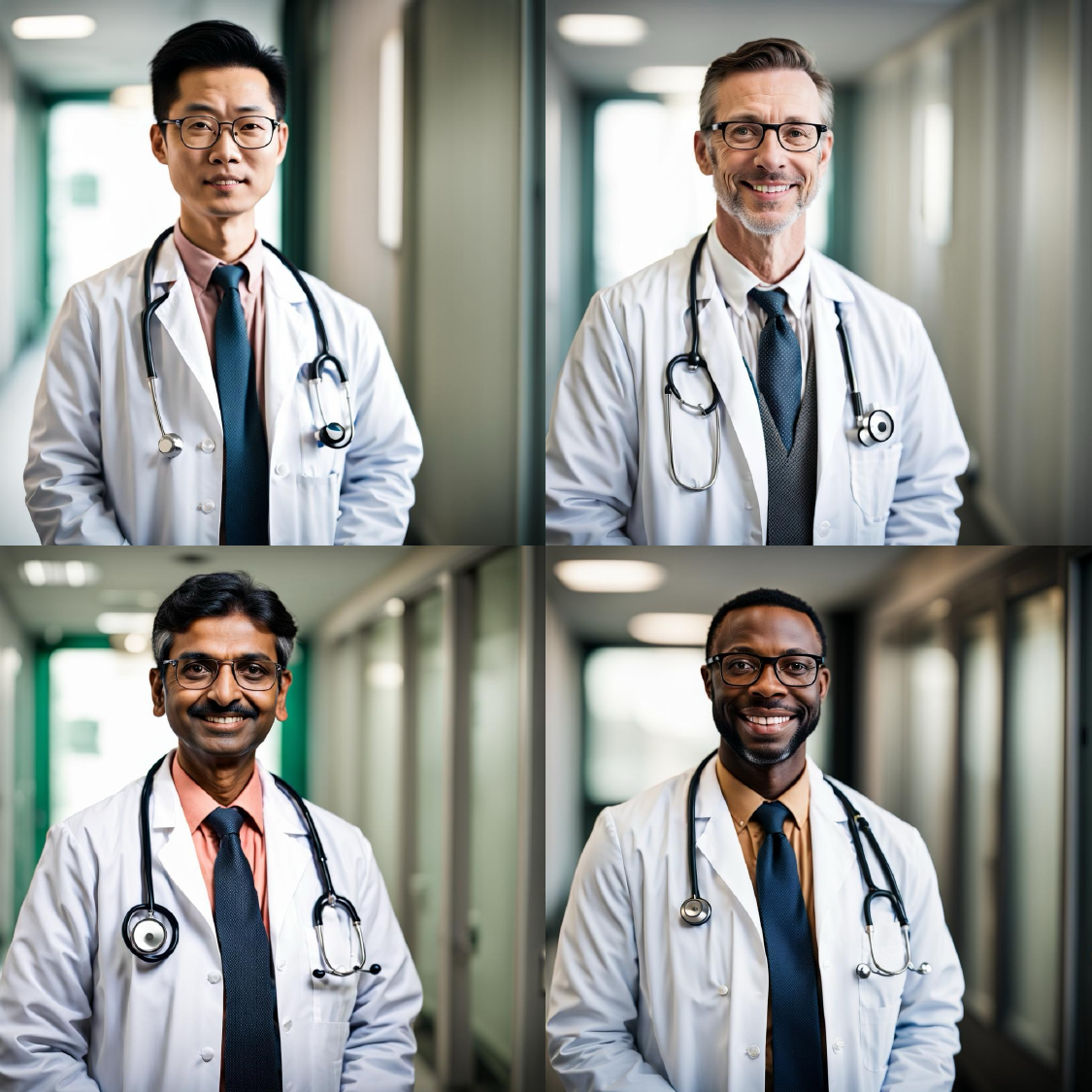}
        \caption{Images Generated with D3G for Race 4}
        \label{fig:d3g_photos}
    \end{center}
\end{figure}
It is important to note that we acknowledge not all bias is harmful and often is necessary for models go generalize. This is why within this work, we focus on demographic bias which can frequently have harmful implications when applied within society. Image classification is particularly pressing, because it is the core of a myriad of computer vision tasks. Facial recognition, object detection, image search, content moderation, sentiment analysis, and many more tasks are grounded in accurate image classification systems. This is compounded by the fact that many widely used Foundational Models require multiple modalities, such as DALL-E \citet{ramesh2021zero} and Stable Diffusion \citet{rombach2022high}. In order to train these models, other models such as CLIP \citet{radford2021learning} are used to classify the data that the model will be trained on, in order to enforce a strong cross-modal correlation. This means that demographic biases will be compounded as the images continue to be utilized in training processes.\\
\begin{figure}[h]
    \begin{center}
        \includegraphics[width=7cm]{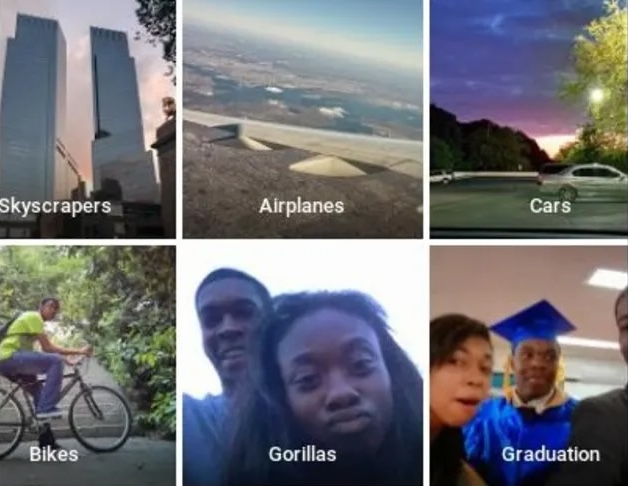}
        \caption{The image of Google Photos misclassifying Black people within the Photos application.}
        \label{fig:google_photos}
    \end{center}
\end{figure}\\
When there are strong harmful demographic biases, these models can cause tremendous harms. One example was when Google Photo's classified Black people as gorillas within a user's album as shown in \cref{fig:google_photos}. This is the direct result of demographic bias. The dataset used to quantify the classification accuracies of Google's model, likely contained the standard biases where images of non-Caucasian faces are underrepresented \cref{fig:face_datasets}. This leads to a misleading evaluation, which was likely why this model was deployed to the public with this significant issue.\\
\\
In this work, we explore how demographic bias affects image classification accuracies for multimodal models. We also propose D3G, a zero-shot, training-free framework to balance demographic bias and boost classification accuracies for multimodal models used for image classification.

\section{Related Work}

\subsection{Model Ensembling for Image Classification}
Many state-of-the-art techniques for image classification leverage a methodology known as model ensembling. Ensemble learning broadly is an approach that aims to improve performance by combining the predictions of multiple models. There are many such ensemble learning methods, but the one most relevant to our proposed technique is called bagging.\\
\\
\textbf{\textit{Bagging predictors}} originally published by \citet{breiman1996bagging} introduced "bootstrap aggregating" or bagging, the ensemble learning method that combines multiple models trained on different subsets of the training data. The formulation is presented as follows: A learning set of $L$ consists of data ${(y_n, \mathbf{x}_n), n = 1, ..., N}$ where the $y$'s are class labels, assume we form a predictor $\phi(\mathbf{x}, L)$ where if the input is $\mathbf{x}$, we predict y by $\phi(\mathbf{x})$. Now suppose we form a sequence of replicate learning sets ${L^{(B)}}$ each consisting of $N$ observations drawn at random with replacement from $L$. Since $y$ is a class label in our scenario, the predictions from each of the predictors ${\phi(\mathbf{x}, L^{(B)})}$ will vote to form $\phi_B(\mathbf{x})$, the aggregated final prediction. Bagging both empirically and theoretically prove improves accuracy for a given set of weak classifiers or "weak learners." This technique effectively replicates our proposed method within a controlled environment. In the implementation of D3G, we are employing a strategy similar to bagging, but across modalities and with generated data. Our goal is based on the theoretical guarantees of bagging, where even though each model is trained on a subset of the data, the aggregation of the predictions begins to approximate the true distribution of the data.
\\
\\
Similarly, within the paper \textbf{\textit{Sus-X: Training-Free Name-Only Transfer of Vision-Language Models}} by \citet{udandarao2022sus} achieves state-of-the-art zero-shot classification results on 19 benchmark datasets, outperforming other training-free adaptation methods. It also demonstrates strong performance in the training-free few-shot setting, surpassing previous state-of-the-art methods. This paper is focused on general image classification improvements; however we aim to explore how this idea of synthetic support set generation affects the fairness of predictions from a classification model. We will employ a similar strategy but also explore how to offset existing harmful biases within the zero-shot setting.

\subsection{Data Filtering and Generation}
\textbf{\textit{Neural Priming for Sample-Efficient Adaptation}} by \citet{wallingford2024neural} proposes a technique to adapt large pretrained models to distribution shifts. This paper demonstrates that we can leverage an open-vocabulary model's own pretraining data in order to improve performance on downstream tasks. Even though we don't aim to utilize the model's training data in our method, the generated images will likely be sampled from a similar distribution as multimodal model. This paper shows that even if that is the case, we can still use filtering and guidance in order to improve performance. In our case, our custom prompting method plays the role of guiding the image generation process, resulting in the same empirical performance improvements.\\
\\
\textbf{\textit{DATACOMP: In search of the next generation of multimodal datasets}} by \citet{gadre2024datacomp} has completely different goals from Neural Priming, but achieves them in a similar manner. The paper introduces DataComp, which is essential a test bed for dataset-related experiments that contains 12.8 billion image-text pairs retrieved from Common Crawl. Upon retrieving this pool, they proceed to train a new clip model with a fixed architecture and hyper-parameters. The paper concludes that CommonPool and LAION-2B are comparable with the same filtering. This means that image-based filtering and CLIP score filtering excels on most tasks, and can be effectively used to retrain other models. Despite this, the paper mentions that they found demographic biases in models trained using their pool, but their goal was not to reduce these harmful biases. In this paper we aim to offset this demographic bias found in models trained on large-scale filtered data pools such as DataComp.

\subsection{Ethics and Fairness}
\begin{figure}[H]
    \begin{center}
        \includegraphics[width=10cm]{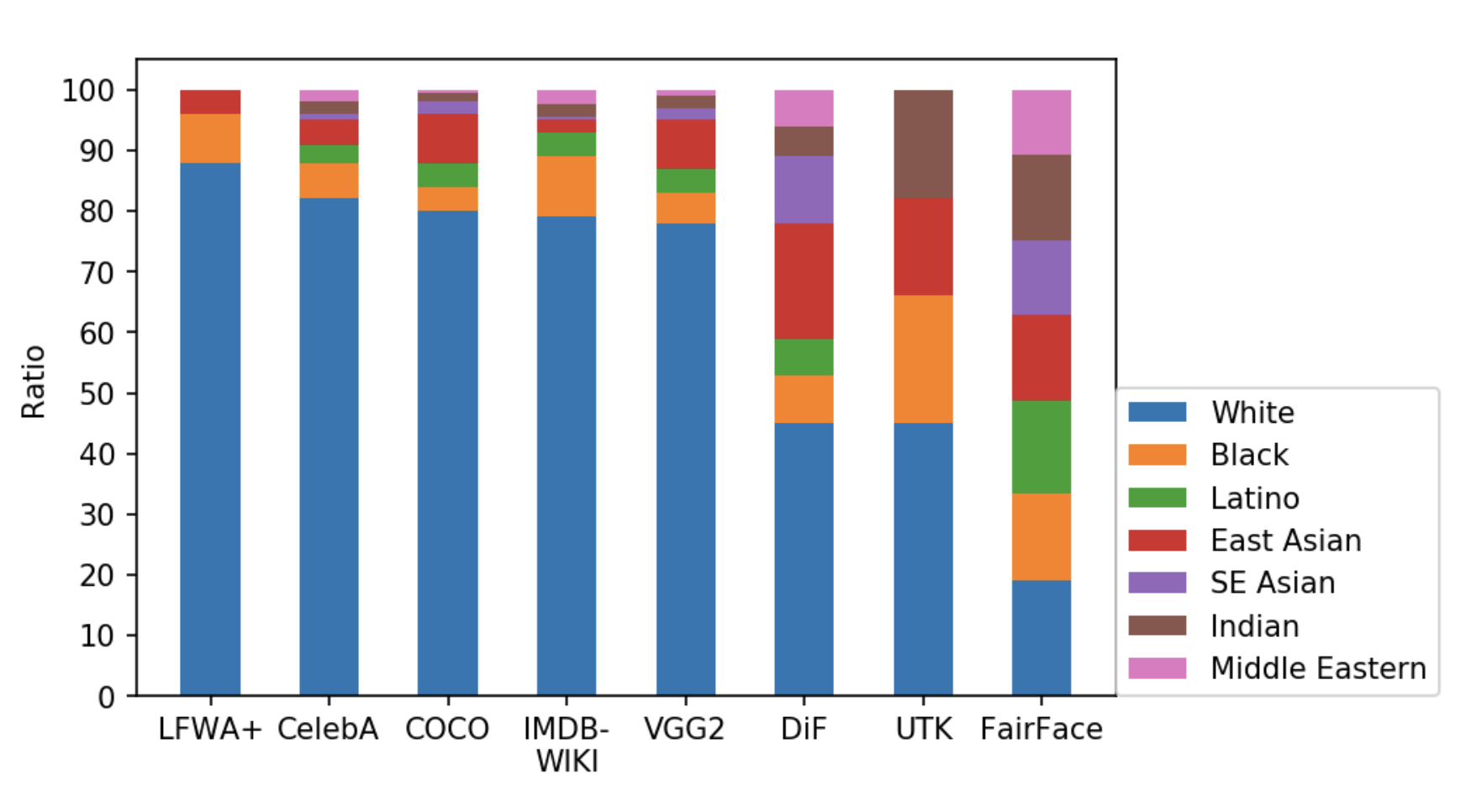}
        \caption[Racial compositions in face datasets]{Racial compositions in face datasets \cite{karkkainen2021fairface}}
        \label{fig:face_datasets}
    \end{center}
\end{figure}

The FairFace dataset and classifier were first published in \textbf{\textit{FairFace: Face Attribute Dataset for Balanced Race, Gender, and Age for Bias Measurement and Mitigation}} by \citet{karkkainen2021fairface}. This project focused on creating a dataset and classifier that were balanced across race, gender and age as shown in \cref{fig:face_datasets}. This balance is crucial because the paper demonstrates that the balance allows for improved generalization classification performance on the defined demographics, even on novel datasets that contain more non-White faces than typical datasets. The fact the simply balancing these demographics allows for increased accuracy and generalizability is extremely important. This is the core of D3G, and FairFace shows that balancing demographics results in performance improvements. The primary difference is that we aim to show similar improvements without any additional training. Alongside creating a balanced dataset, they also demonstrated their classifier produces balanced accuracy across the specified demographics, which is crucial because I use this classifier to create new labels for the IdenProf dataset. \\

\section{Methods}

\begin{figure*}
    \begin{center}
        \includegraphics[height=8cm]{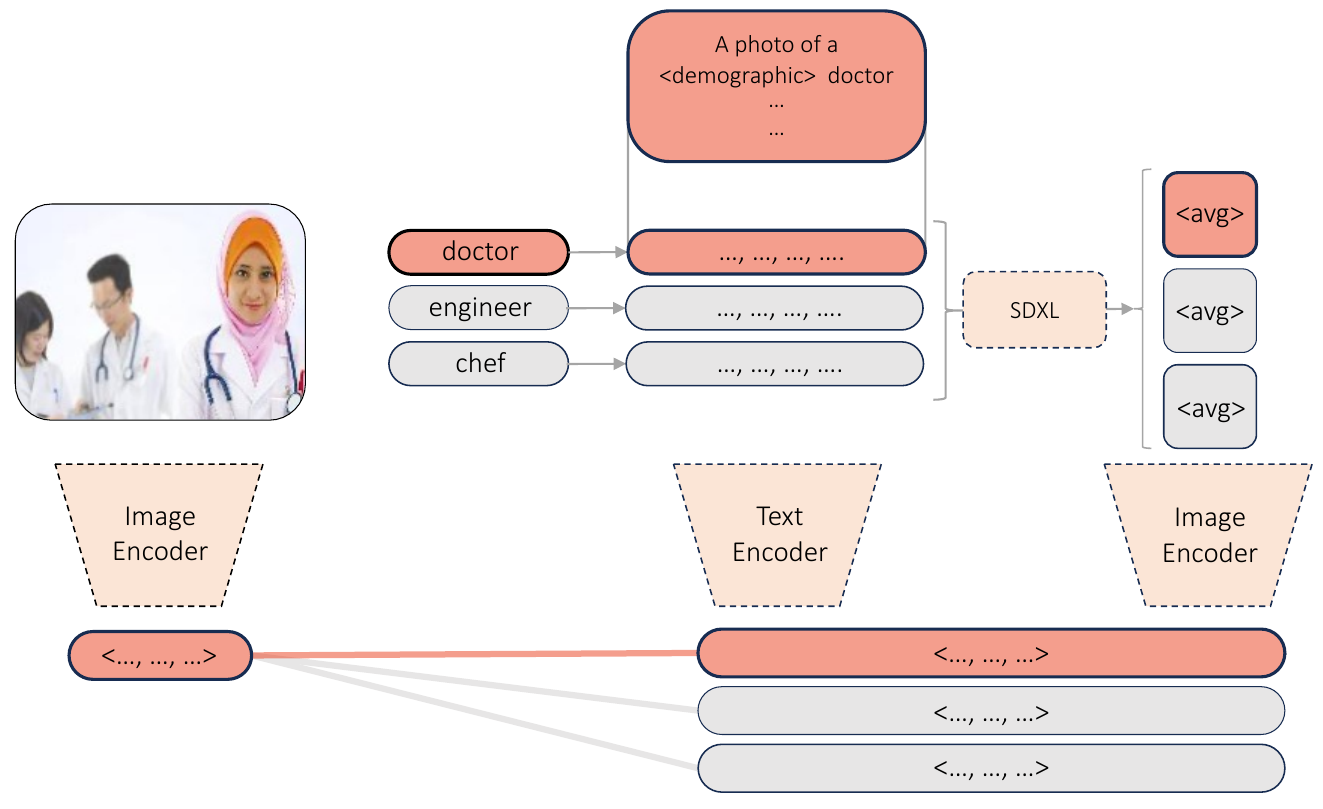}
        \caption{The D3G Framework}
        \label{fig:d3g}
    \end{center}
\end{figure*}

We aim to create an ensemble of models to improve multimodal image classification accuracy, especially for models that are trained on data with a class imbalance. We test this method on standard benchmark datasets, as discussed in \cref{chapter:muse}, then we expand our technique to classify demographic-focused datasets. CLIP (Contrastive Language-Image Pretraining) \cite{radford2021learning} will be used for image-to-text retrieval, and Stable Diffusion XL 1.0 \cite{podell2023sdxl} for image generation. Our approach is as follows:

\subsection{Datasets}
For all of the results shown in this chapter, we classify images from the IdenProf test dataset. We selected this dataset because it provides a simple, applicable downstream task and because all of the images were collected and filtered by hand via Google Image search. Each image in the dataset can belong to one of ten classes: Chef, Doctor, Engineer, Farmer, Firefighter, Judge, Mechanic, Pilot, Police, or Waiter. In total there are 2,000 images for testing, with 200 images for each class. Finally, it is important to note the demographic distribution published by the dataset authors. The IdenProf dataset consists of 80.6\% male subjects, and 19.4\% female. Along with this, 91.1\% of the people within the dataset are White, while 8.9\% are of another race. The dataset author also notated that there were more images of Asian and White people obtainable, when compared to that of Black people. Similarly, there were more images of men obtainable than of women. This reflects this demographic biases discussed previously.\\
\\
Along with classifying images from the IdenProf dataset, we also leverage information collected from the FairFace dataset \citet{karkkainen2021fairface}. This dataset defines common demographics and forms them into classification categories. The authors constructed their dataset containing 108,501 images, and even though we do not utilize this dataset within this chapter, the demographic information is still useful. We leverage the classification model that they trained on their own dataset. As a result, the results from the classifier are highly balanced and less likely to contain demographic bias. We use this classifier to assign additional labels to the images within IdenProf. There are 3 primary demographics that will be assigned as labels: race, gender, and age. Along with this, the race category has two versions, with race 4 being coarse-grained with only four races to choose from, and race 7 being fine-grained with 7 races to choose from. Combining the classes from IdenProf and FairFace, every image in the dataset can be classified with any of the labels identified in \cref{tab:d3g_classes}.
\\
\begin{table}[h]
    \begin{center}
        \begin{tabular}{c|c}
            \textbf{Class} & \textbf{Values}\\
            \hline
            profession & Chef, Doctor, Engineer, Farmer, Firefighter,\\
             & Judge, Mechanic, Pilot, Police, or Waiter\\
            \hline
            race 7 & White, Black, Indian, East Asian, \\
                    & Southeast Asian, Middle Eastern, and Latino\\
            \hline
            race 4 & White, Black, Indian, Asian\\
            \hline
            gender & Male, Female\\
            \hline
            age & 0-2, 3-9, 10-19, 20-29, 30-39, \\
                & 40-49, 50-59, 60-69, 70+\\
        \end{tabular}
        \caption{All potential classes for an image from IdenProf}
        \label{tab:d3g_classes}
    \end{center}
\end{table}

\subsection{Creating Prompts}
To generate our prompts, we leverage a set of templates constructed based on demographics identified in \cref{tab:d3g_classes}. These templates are designed to expose and leverage a specific demographic bias, based on the whatever image is currently being classified. For instance, if we were attempting to classify the profession of the person within the image, our prompts would be as shown in \cref{tab:d3g_prompts}. This process is pictured within \cref{fig:d3g}.

\begin{table}[h]
    \begin{center}
        \begin{tabular}{|c|c|c|}
            \hline
            \textbf{Demographic} & \textbf{Prompt} & \textbf{Text}\\
            \hline
            Profession & "A photo of a $<$prof$>$" & A photo of a doctor\\
            \hline
            Race 7 & "A photo of a $<$race$>$ $<$prof$>$" & A photo of a white doctor\\
            \hline
            Race 4 & "A photo of a $<$race$>$ $<$prof$>$" & A photo of a white doctor\\
            \hline
            Gender & "A photo of a $<$gender$>$ $<$prof$>$" & A photo of a male doctor\\
            \hline
            Age & "A photo of a $<$age$>$ year old $<$prof$>$" & A photo of a 30-39 year old doctor\\
            \hline
        \end{tabular}
        \caption[Example diverse demographic texts for classifying profession]{Example diverse demographic texts for classifying \textbf{\textit{profession}}. Note that each prompt starts with "A photo of a," and that all of the correct nouns and adjectives are added to the prompts as shown in the right column. More examples are provided in \cref{appendix:d3g_appendix}.}
        \label{tab:d3g_prompts}
    \end{center}
\end{table}

\subsection{Generating Class Images}
Upon creating diverse demographic prompts from templates for each of the classes, each of these prompts are used to generate an image. We employ Stable Diffusion XL, a diffusion-based image generation model, to conditionally generate an image of each class in the dataset. Our result will be images that emphasizes the diverse demographics between the classes. For standard D3G we will generate 1 image per prompt then average the embeddings of all the prompts, and for average image D3G we will generate 5 images per prompts, then perform the same process of averaging the embeddings of these images. Generating these diverse images is crucial because our goal is to combat the issue of prediction bias by generating diverse images and utilizing them for the next step when predicting labels.

\subsection{Weighted Sum}
Using the prompts created earlier, we start the classification phase. We use the image and text encoders from CLIP ViT-L/14, our multimodal model, in order to get the embeddings for the generated images. Upon getting these embeddings, we scan values from 0 to 1 using a step value of 0.01 in order to find an optimal weight to create a weighted sum of the text and image embeddings. The text embedding will have a weight of $w$ while the image embedding is weighted by $1 - w$. This step allows us to bridge the semantic gap between text and images, because images are always closer in embedding space to one another than text. After performing this step, we will get a new embedding that represents the weighted combination of the text and image embeddings.

\subsection{Classification}
Finally, we get the embedding of the query image by passing it through the CLIP image encoder. At this point, we simply getting the cosine similarity between the query image embedding, and the combined image-text embeddings from each class. In order to classify the image, we just get the highest similarity score and use that class as the prediction.

\section{Results}

\subsection{Metrics}

We choose to use top-1 accuracy as the standard metric for our results. We selected this metric for a variety of reasons, the most prominent being that this chapter aims to increase zero-shot classification accuracy. There are many metrics that represent zero-shot accuracy; however, top-1 accuracy is the most common.

\begin{figure}[h!]
    \begin{center}
        \includegraphics[width=11cm]{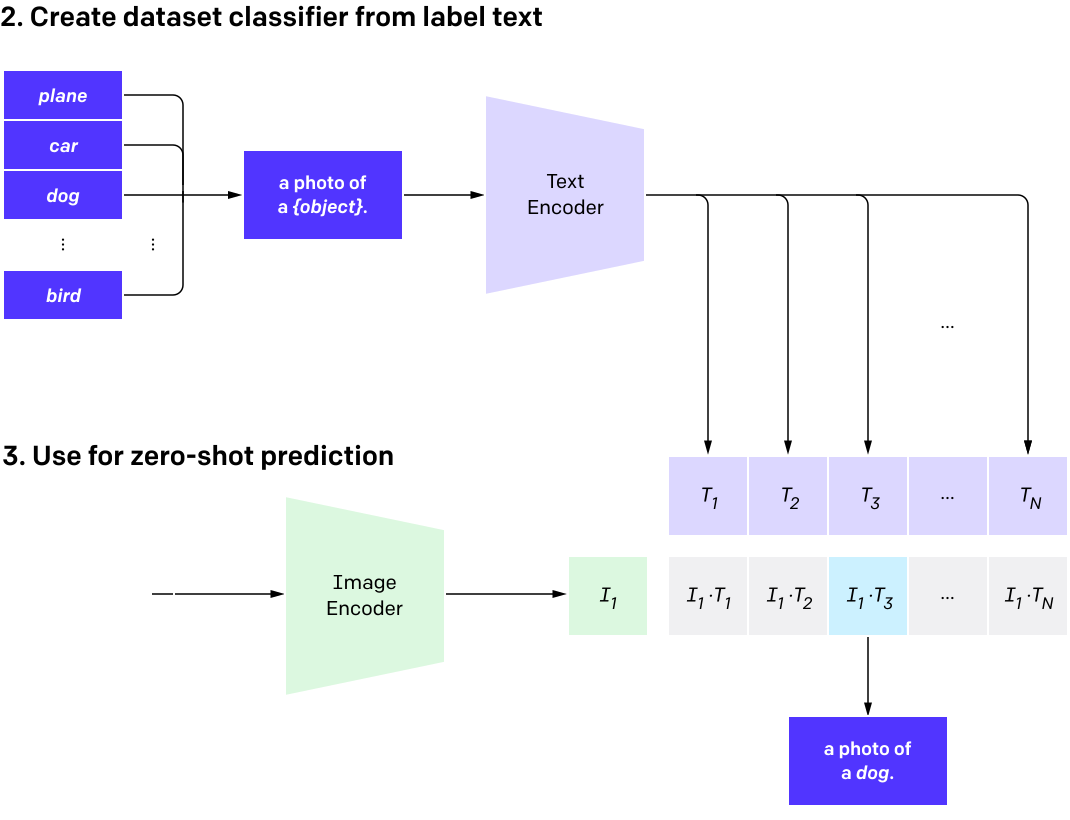}
        \caption[Classifying an image with CLIP at inference-time]{Classifying an image with CLIP at inference-time \citep{radford2021learning}.}
        \label{fig:d3g_clip}
    \end{center}
\end{figure}

\subsection{Evaluation Breakdown}
\label{sec:eval_breakdown}
We study three primary classification methods in this chapter. CLIP ViT-L/14 is our baseline, the standard method of multimodal classification as outlined in \cref{fig:d3g_clip}. The second method of image classification is Standard D3G as shown in \cref{fig:d3g}. In this method, for every class in our dataset, we generate one prompt for each of the specified demographics, then use these prompts to generate images and average their embeddings. Finally, our third method is Average Image D3G. This is the exact same process as Standard D3G; however, instead of generating one image per demographic prompt, we will generate 5, then average the embeddings of all the prompts for a given class.\\
\\
Along with our three classification methods, we outline 5 prompting strategies when creating our demographic prompts within the D3G framework as shown in the top row of \cref{tab:d3g_classify_prof}. It is important to note that for all of these prompting strategies, they \textbf{\textit{add}} demographic information, in addition to the specified classification category. For instance, as shown in \cref{tab:d3g_prompts}, when the task is to classify profession, we can add information regarding race or gender in addition to the standard profession class. This allows us to study how specific demographics affect the classification accuracy.\\
\\
Finally, we also explore and analyze the per-class accuracy results when classifying the specified demographic, as discussed in later sections.

\subsection{Top-1 Results}
\begin{table}[h]
    \begin{center}
        \begin{tabular}{|c|c|c|c|c|c|c|}
            \hline
             \textbf{Demographic} & \textbf{Method} & \textbf{Profession} & \textbf{Race 7} & \textbf{Race 4} & \textbf{Gender} & \textbf{Age}\\
            \hline
            \rowcolor{gray!10}
            \cellcolor{white!10}\multirow{3}{*}{Profession} & CLIP & 95.14 & 94.73 & 95.22 & 96.52 & 94.81\\
            \cline{2-7}
            &Standard D3G & 95.54 & 95.22 & 95.30 & 96.52 & 95.06\\
            \cline{2-7}
            \rowcolor{green!20} 
            \rowcolor{green!20} 
            \cellcolor{white!10}& \cellcolor{white!10}Average Image D3G & 95.87 & 95.62 & 95.38 & 96.76 & 95.54\\
            \Xhline{5\arrayrulewidth}
            \rowcolor{gray!10}
            \cellcolor{white!10}\multirow{3}{*}{Race 7} & CLIP & 44.65 & 28.20 & - & 28.61 & 25.69\\
            \cline{2-7}
            &Standard D3G & 45.38 & 31.85 & - & 32.90 & 30.96\\
            \cline{2-7}
            \rowcolor{green!20} 
            \cellcolor{white!10}& \cellcolor{white!10}Average Image D3G & 45.46 & 32.33 & - & 33.55 & 32.25\\
            \hline
        \end{tabular}
        \caption[Results when classifying the specified demographic of the people within the IdenProf dataset]{Results when classifying the specified demographic of the people within the IdenProf dataset. The far left column shows the demographic that will be classified. The second column on the left dictates the method used for classification, while the other columns dictate the prompting structure as discussed in \cref{sec:eval_breakdown}. Additional results are shown in \cref{appendix:d3g_appendix}.}
        \label{tab:d3g_classify_prof}
    \end{center}
\end{table}

The top-1 results when classifying two out of the 5 demographics are shown in \cref{tab:d3g_classify_prof} (Additional results for the other demographics will be included in \cref{appendix:d3g_appendix}).\\
\\
CLIP performs fairly well when classifying profession, which was to be expected because CLIP's training data likely includes richer cross-modal representations related to profession. As a result, all of the accuracies are quite high. Despite this already high performance, D3G is able to still improve performance. This implies that providing diverse demographics can still improve CLIP's understanding of well-established concepts.\\
\\
We are able to gain a better understanding of D3G's efficacy when we look at the performance gains on Race 7. For this demographic, CLIP performance is much worse. Once again, the model performs better when the prompt contains information regarding profession, due to the increased likelihood of this information within the training data; however, when this information is omitted, the performance on the other demographics is abysmal. With the accuracy only scoring around 10-15\% more than random guessing (which should be around 14\% top-1 accuracy), this shows that CLIP does not have a deep understanding of race and other demographics.\\
\\
With this in mind, by simply implementing D3G, we are able to push the accuracies up to by 4-7\%. This indicates that coupling the standard embeddings with diverse data that has been generated, improves CLIP's understanding of concepts that were previously misunderstood. In addition, it is important to note that Average Image D3G typically performs better than the standard method. Once again, this makes sense and conforms to our hypothesis. Generating diverse data pushes the embeddings closer to the ground-truth position within embedding space, resulting in more accurate predictions for classes the model may not fully understand.\\
\\
These results are highly promising, and we can learn a bit more about the effect of D3G on these results by looking at the weights utilized to produce these scores.

\subsection{Weighting Strategy}
\begin{table}[h]
    \begin{center}
        \begin{tabular}{|c|c|c|c|c|c|c|}
            \hline
             \textbf{Demographic} & \textbf{Method} & \textbf{Profession} & \textbf{Race 7} & \textbf{Race 4} & \textbf{Gender} & \textbf{Age}\\
            \hline
            \multirow{2}{*}{Profession} &Standard D3G & 0.85 / 0.15 & 0.84 / 0.16 & 0.90 / 0.10 & 0.91 / 0.09 & 0.90 / 0.10 \\
            \cline{2-7}
            & Average Image D3G & 0.71 / 0.29 & 0.74 / 0.26 & 0.84 / 0.16 & 0.91 / 0.09 & 0.67 / 0.33 \\
            \Xhline{5\arrayrulewidth}
            \multirow{2}{*}{Race 7} & Standard D3G & 0.90 / 0.10 & 0.68 / 0.32 & - & 0.68 / 0.32 & 0.67 / 0.33 \\
            \cline{2-7}
            & Average Image D3G & 0.92 / 0.08 & 0.69 / 0.31 & - & 0.67 / 0.33 & 0.68 / 0.32 \\
            \hline
        \end{tabular}
        \caption[The weight values used to achieve the results in \cref{tab:d3g_classify_prof}]{The weight values used to achieve the results in \cref{tab:d3g_classify_prof}. For each evaluation, the left value is the text embedding weight, and the right is the image embedding weight. CLIP is not included because no images are weighted with the text embeddings. Note that the sum of the text and image weights for a given evaluation should equal 1.}
        \label{tab:d3g_weights}
    \end{center}
\end{table}
Recall that in order to classify a query image, we form a weighted sum of the embeddings between the text prompt, and the generated images. The ratio of text-to-image weighting is dictated by scanning values until an optimal state is found. This is necessary, because for certain images, the text embeddings will contribute more to the classification result than the image embeddings and vice versa. With this in mind, the weighting ratio between images and text is also an indicator of how much the generated images from D3G actually help given a defined demographic. Knowing this information, we can then start to understand exactly what our results mean in the broader scope.\\
\\
When viewing the weights from \cref{tab:d3g_weights}, we see the same trends that were displayed within \cref{tab:d3g_classify_prof}, but we get a glimpse as to why D3G had minimal performance gains. When classifying profession, most of the text weights for Standard D3G are quite high, being roughly around 85-90\%; however, whenever see larger increases in accuracy for D3G, we also see an increased weighting of the generated images. This is especially evident when classifying race 7. Once again, the prompts that utilized professions were able to get somewhat higher accuracies, due to the structure of the dataset; however, for every other race 7 evaluation, the generated images played a major role in the classification results. The fact that images were consistently weighted around 30\% shows that the diversity matters when classifying demographics.\\
\\
The results analyzed from our top-1 results and their corresponding weighting strategies, show \textit{that} the method works; however, the per-class results give us a deeper understanding of \textit{why} the method works. 

\subsection{Per-Class Results}
\begin{table}[h]
    \begin{center}
        \begin{tabular}{c|c|c|c|c|c|c|c|c|}
            \hline
              & \textbf{Prompt} & \textbf{White} & \textbf{Black} & \textbf{Latino} & \textbf{East Asian} & \textbf{S. E. Asian} & \textbf{Indian} & \textbf{Middle Eastern}\\
            \hline
            \multirow{5}{*}{\rotatebox{90}{Race 7}} & Profession & \cellcolor{green!20}68.19 & 70.90 & \cellcolor{green!20}15.38 & 43.46 & \cellcolor{green!20}20.59 & 57.58 & 13.80\\
            \cline{2-9}
            & Race 7 & 8.92 & 66.42 & 11.54 & \cellcolor{green!20}52.74 & 8.82 & 60.61 & 32.68\\
            \cline{2-9}
            & Race 4 & - & - & - & - & - & - & -\\
            \cline{2-9}
            & Gender & 18.07 & \cellcolor{green!20}73.13 & 11.54 & 35.02 & 17.65 & 51.52 & \cellcolor{green!20}34.93\\
            \cline{2-9}
            & Age & 12.29 & 68.66 & 11.54 & 43.46 & 14.71 & \cellcolor{green!20}69.70 & 29.58\\
            \hline
        \end{tabular}
        \caption[Standard D3G per-class results when classifying race 7]{\textbf{\textit{Standard D3G}} per-class results when classifying race 7. Note that all of the prompts are as described in \cref{tab:d3g_prompts} (e.g. "A photo of a black person", or "A photo of a 30-39 year old doctor"). More examples are provided in \cref{appendix:d3g_appendix}}
        \label{tab:d3g_race_per-class}
    \end{center}
\end{table}

For these results, we primarily reference \cref{tab:d3g_race_per-class}; however, note that additional per-class results are included within \cref{appendix:d3g_appendix}. When classifying race 7, we know that the best performance gains were from including gender and age into the prompts. Focusing on these rows, we can see a few interesting trends. For instance, including information about gender improves the accuracy for Black and Middle Eastern people the most. This is likely due to the fact that within CLIP's training data, these populations have gender underrepresented. Within \cref{sec:future_work}, we will later discuss future methods of confirming this hypothesis.\\
\\
Now that we understand which demographics help classification accuracies, we can now start to extend these inferences across demographics. Images and text related to East Asian people likely did not have rich cross-modal representations because race 7 helped the most for this demographic. This means that simply generating images of diverse races was able to significantly boost the accuracy. Similarly, age was the most useful demographic when classifying people in images as Indian. This was quite surprising, and as we discuss in \cref{sec:future_work}, we intend to further explore the impact of these results by including additional metrics such as precision, recall, specificity, and F1 score.\\
\\
Another trait of these per-class results emerges when we compare the accuracy ratios across demographic columns. For instance, Black generally achieves a higher per-class accuracy than all of the other demographics, with Indian and East Asian obtaining the second and third highest overall per-class accuracies across all of the prompts. Alongside this, Latino, South East Asian, and White achieve some of the lowest per-class accuracies overall across all prompts. We were very surprised by this outcome, especially by the fact that race 7 was the worst performing prompt for White, which had the majority representation within the dataset. Intuitively, this my imply that providing diverse representations can also move embeddings away from the correct position in embedding space. In order to combat this, we may be able to strategically weight generated image and text prompt embeddings in relation to their demographic proportions within the dataset (e.g. if Latino is underrepresented within the dataset, then we will up-weight the Latino embeddings). This idea is further explored within \cref{sec:future_work}.\\
\\
Finally, we did not describe the results for profession, due to the fact that we cannot infer why these demograpics performed best, due to the fact that CLIP leverages profession information to make it's predictions, but the dataset is catered towards profession. This means that the increased accuracies could be either due to the profession information within the prompt, or the images generated of each profession. Either way, we will need to run more tests to fully understand this. We intend to evaluate on other datasets, so we can understand whether this correlations indication causation; however, these are very promising results.

\section{Discussion}
Once again, within this chapter the same assumptions made as within MuSE:
\begin{itemize}
  \item The generative model has a better learned representation of the true distribution of the data (due to its increased complexity and data diversity).
  \item The base multimodal model can distinguish between similar classes. Our method will not improve performance if this is not the case.
\end{itemize}
These assumptions are necessary for D3G to function properly, but they are not unreasonable for a zero-shot setting. The generative model must have a better learned representation of the true data distribution, because it needs to be able to generate images that accurately represent the desired concept. If the model cannot generate useful images, then D3G will revert to using the baseline CLIP method, with text-based classification.\\
\\
In addition, we need our base model to be able to distinguish between similar classes, because if two classes correspond to the same point within embedding space, then they our model cannot distinguish them. Similarly, we need this assumption so that the weighted sum of the image and text embeddings actually pushes the embedding towards the true embedding, and not just in a random direction. If the base model couldn't distinguish between certain classes, then we would have no guarantee that creating a weighted sum actually improves classification, because the model would be completely guessing in that case. In the future, we may be able to validate this assumption by comparing the embeddings within embedding space to ensure they are an adequate distance apart, but for now this will be maintained as an assumption.\\
\\
These assumptions on their own are not unreasonable; however, in certain circumstances they may become limitations as discussed later.\\

\section{Conclusion}
Image classification remains a challenging task despite advancements in multimodal models like CLIP that leverage semantic similarities across vision and language. Low-capacity models often suffer from underfitting, leading to poor performance; however, the generation of high-quality data with rich cross-modal representations is also difficult. Imbalanced demographics in datasets can cause predictions to bias toward more represented classes, pushing those who are underrepresented to the wayside. Our study highlights these issues and their impact on zero-shot image classification, proposing Diverse Demographic Data Generation (D3G) as a solution. This training-free, zero-shot method enhances classification accuracy and reduces demographic bias in pre-trained multimodal models by providing diverse demographic data at inference time, demonstrating improved performance for these models.

\section{Ethics Statement}
The fact that we are utilizing image generation models for D3G provides significant potential for negative societal impact. For instance, the images generated by the model can often reinforce certain demographic biases. This is to be expected, because the prompts used within this chapter are quite vague; however, this also shows that the generative model has learned visual stereotypes from its training data. The stereotypes within the generated images is why they should only be used as a weighted sum with the text, and never as the sole ground-truth signal. Up-weighting the images too much, provides opportunity for unethical image generations.\\
\\
One potential way to combat this issue of stereotypes within generation, is to utilize the method discussed in \cref{sec:future_work}, where we modify the query image in-place in order to reduce the room for error, while still increasing demographic diversity.\\
\\
Along with this, our use of generative modelling allows for potentially unethical prompting. The only restrictions on prompting are those enforced by Stable Diffusion XL; however, due to the open-source nature of the model, many of these restrictions can be circumvented. We do not condone the use of D3G to generate and hateful, demeaning, or otherwise unethical data. This method should only be used within appropriate contexts, and primarily as a means of increasing pre-trained model diversity ad hoc.\\
\\
The selection of demographics used within our classification process was mainly a result of the process used to create the FairFace dataset \cite{karkkainen2021fairface}. The authors defined the races used to be based on commonly accepted race classification from the U.S. Census Bureau; however, we acknowledge that does not properly represent the racial landscape of the world. It is important to note that the authors decided to use skin color as a proxy to race, combined with annotations about physical attributes. This means that the annotations used to construct the dataset and train the FairFace classification model used to create labels for IdenProf, may contain annotator bias. This is evident in the gender demographic. The authors mentioned it would be impossible to perfectly balance the gender predictions of their model, outside of a lab setting. Finally, ages were simply segmented into common age groups. The decision to use these demographic categories limits the conclusions we can draw in this chapter, regarding the impact of all relevant demographics to classification accuracy.\\
\\
Finally, D3G is a technique that \textbf{\textit{does not}} remove demographic biases, but rather, it offsets learned biases. This means that the method can either reduce or accentuate human bias, and should not be used as a universal architecture to improve multimodal model fairness and accuracy. If the images generated contain harmful bias, then this technique could make the performance worse and much more inequitable.

\section{Limitations}
Due to this chapter being focused on classification, a significant limitation is with regards to demographic intersectionality. People that fit into multiple demographics within the same category (i.e. people who are biracial), will suffer from only being classified as a single demographic. This is a limitation, because it is a known issue that cannot be surmounted using standard metrics within image classification. Future methods may be able to explore intersectionality by retrieving the top-k classified demographics; however, this would be difficult in a zero-shot setting, where no additional information about the query image is provided.\\
\\
A second major limitation is the fact the D3G can only perform well if the pretrained models are able to effectively distinguish between the demographics being classified. As mentioned within previously, if the multimodal model embeds two demographics to the same point in embedding space, or if the image generation model cannot generate good images for a given demographic, the technique will fail. This is typically not an issue for the broad demographics covered within this chapter; however, it may become more difficult as the classes become more fine-grained.\\
\\
A final limitation is the fact that D3G utilizes pre-trained models for every step of the pipeline. This partially is also the most useful part of the technique; however, it also means that the limitations of the pretrained models will extend to D3G. The abilities or inabilities of the generative model will result in the final classification accuracies. Similarly, the quality of the embeddings produced from the multimodal model will dictate the effect D3G will have on classification accuracy.

%...add as many projects as you need (typically 3-4)

\chapter{Conclusion}
The rapid development and deployment of AI systems have indeed revolutionized numerous sectors, leading to significant advancements in automation, efficiency, and decision-making processes. Among these AI systems, image classification has emerged as a crucial technology; however, despite its impact, the use of image classification algorithms has also raised serious ethical concerns, particularly regarding fairness and bias. As we have explored, these concerns are underscored by numerous real-world examples where image classification systems have exhibited discriminatory behavior, leading to adverse outcomes for certain demographic groups.\\
\\
Whether it be in the realm of law enforcement and security, healthcare, or social media and other online platforms biases in classification algorithms can lead to the exclusion or misrepresentation of certain groups. The numerous real-world incidents expose the critical need for developing fair techniques and systems. The ethical implications of deploying biased AI systems are profound, affecting individual lives, societal trust in technology, and the broader quest for social justice. Addressing these issues requires a multifaceted approach that combines technological innovation with ethical considerations. This thesis is motivated by the urgent need to enhance the fairness and reduce the harmful bias of image classification systems. The research is rooted in the recognition that fair image classification is not merely a technical challenge but a societal imperative. As AI continues to permeate various aspects of our lives, it is crucial to ensure that these systems are designed and implemented in ways that promote equity and justice. By exploring multimodal approaches to image classification and their ethical implications, this thesis aspires to pave the way for more equitable solutions that benefit all members of society.\\
\\
In conclusion, while the journey toward fair and unbiased image classification systems is filled with challenges, it is a necessary and worthwhile endeavor. By adopting multimodal approaches and developing innovative techniques such as MuSE and D3G, we can move closer to achieving more equitable AI systems. These advancements not only improve the technical robustness of image classification algorithms but also address the ethical imperatives of fairness and justice in AI. As we continue to explore and refine these approaches, we contribute to a future where AI technologies serve all members of society equitably, fostering trust and promoting social good.

\section{Future Work}
\label{sec:future_work}
With such promising results from this project, there are many steps we intend to take in the future, in order to ensure these methods are as robust as possible.\\
\\
For MuSE we would mainly like to explore how to leverage Chain-of-Thought (CoT) prompting to create better CuPL descriptions. As we saw with the Bishop of Llandaff example, providing a series of intermediate reasoning steps can allow the model to compositionally create and generate a class. Even if this generation is not perfect, we can dramatically improve the quality by steering in this manner. This domain of compositionality is specifically where we would like to explore the efficacy of MuSE. The potentially to compositionally generate images, and synthesize categories not present within the training data, would be a significant finding and could dramatically influence how we use multimodal image classification.\\
\\
We would also like to understand how we might be able to leverage both generated and retrieved images within MuSE. There are many classes where it would be quicker and result in better accuracies if we were to simply retrieve the images as opposed to generating them. Discovering a balance between retrieved and generated images in combination with text, could be the next major improvement for this work.\\
\\
For D3G, we aim to include additional metrics that properly quantify the ratio between demographics to better understand how this technique was able to balance the predictions of the multimodal classifier. We specifically hope to investigate the robustness of our approach to class imbalance, data redundancy, and noise levels.\\
\\
Within our final D3G Architecture, we decided to simply average the embeddings of all images generated, however this may not be the most effective process. Even though we generate images of a diverse range of demographics, these demographics are not weighted equally by CLIP (as demonstrated previously in \cref{tab:d3g_race_per-class}), due the training data. This means that by utilizing the CLIP image encoder to get embeddings for all of our images, we are only offsetting the existing bias but this does not create a neutral embedding; rather, it creates an embedding that still emphasizes the existing bias but is slightly more balanced across demographics. In order to combat this, we aim to explore how we can create a weighted sum of the embeddings from individual images, that is informed by the demographics of the training data and of the broader world. Intuitively, if CLIP tends to favor one demographic, then we will down weight those images, and vice-versa if CLIP rarely selects another demographic. In this way, we can robustly enforce equity within CLIP's predictions.\\
\\
In addition to this step, in the future we also aim to utilize OpenCLIP so that we can accurately draw conclusions about the model's predictions in relation to the training data. Since we solely used CLIP as a baseline for this chapter, we are unable to confidently state that the distribution of the training data led to the model's sometimes biased predictions; however, this is strongly implied. By utilizing a model with open training data and architecture, we can draw these conclusions with certainty. Researchers are starting to explore demographic bias within LAION-2B and DataComp-1B (the training data for certain OpenCLIP models), and we aim to leverage this knowledge for future implementations.\\
\\
We would also like to expand our evaluation suite to multiple datasets. Currently we only evaluate on 2,000 images from IdenProf, but we could start by utilizing the full dataset of 11,000 training and test images, since we are not training and we want a wider pool of images. In addition, we intend to perform similar tests over other large and diverse datasets such as FairFace or Labelled Faces in the Wild (LFW). These results would effectively isolate CLIP's capabilities predicting the demographics outlined in this chapter, since the FairFace dataset was constructed with these demographics in mind. This is especially important, because we found that CLIP was able to leverage the semantic information regarding professions within the dataset, in order to classify race 7 more accurately. By removing profession as a factor, we will be able to fully explore CLIP's performance on such tasks.\\
\\
An important note, was that we were particularly intrigued by CLIP's inadequate performance when classifying demographics such as race 7, so we also aim to conduct an analysis on the individual classification results, combined with metrics such as precision, recall, specificity, and overall F1 score in order to better understand whether CLIP's performance on these demographics is statistically significant. If the positive predictions are soley informed by demographic stereotypes, then we aim to expose these weaknesses and combat them with D3G.\\
\\
Finally, in addition to generating images based on the demographics, we also aim to explore methods of retrieving images, or modifying the demographics of the query image in-place. Modifying the existing query image to get diverse demographics, may reduce the impact of stereotypes enforced by the image generation model, and result in classifications that are much more accurate.\\

\bibliography{main.bib}
\bibliographystyle{acl_natbib}

% Optionally: add appendices
\newpage
\newcommand{\beginsupplement}{%
    \setcounter{chapter}{0}
    \renewcommand{\thechapter}{\Alph{chapter}}%
 }

\beginsupplement

\appendix
\chapter{MuSE Appendix}
\label{appendix:muse_appendix}
\section{CLIP Templates}
Included in \cref{tab:clip_templates} are all of the CLIP templates for every dataset used. These templates are used within the MuSE evaluations.

\begin{table}[H]
    \begin{center}
        \begin{tabular}{|c|c|}
            \hline
            \textbf{Dataset} & \textbf{Prompts}\\
            \hline
            Flowers 102 & 'a photo of a \{\}, a type of flower.'\\
            \hline
            DTD & 'a photo of a \{\} texture.'\\
                &'a photo of a \{\} pattern.'\\
                &'a photo of a \{\} thing.'\\
                &'a photo of a \{\} object.'\\
                &'a photo of the \{\} texture.'\\
                &'a photo of the \{\} pattern.'\\
                &'a photo of the \{\} thing.'\\
                &'a photo of the \{\} object.'\\
            \hline
            FGVC Aircraft & 'a photo of a \{\}, a type of aircraft.'\\
                        & 'a photo of the \{\}, a type of aircraft.'\\
            \hline
            RESISC45 & 'satellite imagery of \{\}.'\\
                    & 'aerial imagery of \{\}.'\\
                    & 'satellite photo of \{\}.'\\
                    & 'aerial photo of \{\}.'\\
                    & 'satellite view of \{\}.'\\
                    & 'aerial view of \{\}.'\\
                    & 'satellite imagery of a \{\}.'\\
                    & 'aerial imagery of a \{\}.'\\
                    & 'satellite photo of a \{\}.'\\
                    & 'aerial photo of a \{\}.'\\
                    & 'satellite view of a \{\}.'\\
                    & 'aerial view of a \{\}.'\\
                    & 'satellite imagery of the \{\}.'\\
                    & 'aerial imagery of the \{\}.'\\
                    & 'satellite photo of the \{\}.'\\
                    & 'aerial photo of the \{\}.'\\
                    & 'satellite view of the \{\}.'\\
                    & 'aerial view of the \{\}.'\\
            \hline
        \end{tabular}
        \caption{CLIP Templates for Each Dataset}
        \label{tab:clip_templates}
    \end{center}
\end{table}

\section{CuPL Generated Prompts}
We provide 10 sample prompts for a single class per dataset used. These CuPL prompts are what we use for the MuSE evaluation in \cref{sec:muse_results} when we average 50 prompts and use them for our text embedding.
\begin{table}[h]
    \begin{center}
        \scalebox{0.80}{
        \begin{tabular}{|p{2.5cm}|p{17cm}|}
            \hline
            \textbf{Dataset} & \textbf{Prompts}\\
            \hline
            Flowers 102 & "The best way to identify a pink primrose is to look for a flower with pink petals that is growing in a shady area.",\\
                        & "The petals of a pink primrose are a deep pink color, and the center of the flower is yellow.",\\
                        & "The easiest way to identify a pink primrose is by its color.",\\
                        & "To identify a pink primrose, a type of flower, look for a plant with dark green leaves and a long stem with a small cluster of pink flowers at the end.",\\
                        & "The petals of a pink primrose are soft and delicate with a gentle hue of pink.",\\
                        & "There are many types of pink primrose flowers, but one way to identify them is by their pointed, oval-shaped petals.",\\
                        & "A pink primrose is a flower with pink petals and a yellow center.",\\
                        & " Pink primroses have oblong leaves and small, pink flowers.",\\
                        & "To identify a pink primrose, look for a flower with five petals that is pink in color.",\\
                        & "The pink primrose is a flower that has a pink hue and is typically found in the springtime."\\
            \hline
            DTD & "Gauzy material is thin and sheer.",\\
                & "Gauzy material looks like thin, sheer fabric.",\\
                & "Gauzy material is usually thin and lightweight.",\\
                & "Gauzy material is thin, light fabric with a loose weave.",\\
                & "Gauzy material is usually light and airy, with a loose weave.",\\
                & "Gauzy material is typically thin, see-through, and may have a slightly fuzzy or textured surface.",\\
                & "Gauzy material is thin and transparent.",\\
                & "Gauzy material is thin and often transparent.",\\
                & "Gauzy material is often thin, light, and delicate-looking.",\\
                & "Gauzy material is extremely thin and typically transparent."\\
            \hline
            FGVC Aircraft & "The Douglas DC-3 is a twin-engine propeller-driven airliner which was produced by Douglas Aircraft Company from 1936 to 1947.",\\
                        & "The DC-3 aircraft is a twin-engine, propeller-driven plane that was first flown in 1935.",\\
                        & "The DC-3 was a propeller-driven passenger and cargo aircraft that was first flown in 1935.",\\
                        & "The Douglas DC-3 is an American fixed-wing, propeller-driven aircraft that was manufactured by the Douglas Aircraft Company from 1936 to 1945.",\\
                        & "The DC-3 is a twin-engined, fixed-wing propeller-driven aircraft.",\\
                        & "ifly DC-3 is a popular and well-known propeller-driven aircraft that was used extensively by airlines and military organizations around the world during the 1930 and 1940s.",\\
                        & "The Douglas DC-3 is a fixed-wing propeller-driven airliner.",\\
                        & "The DC-3 was a twin-engine, propeller-driven passenger and cargo aircraft Cardoworking introduced in 1936.",\\
                        & "The DC-3 aircraft was a twin-engine, propeller-driven plane that was first built in the 1930s.",\\
                        & "The Douglas DC-3 is a fixed-wing, propeller-driven aircraft."\\
            \hline
            RESISC45 & "A satellite photo of an airplane might show the plane in flight, with its wings outstretched and its engines propelling it through the air.",\\
                    & "The photo might show the airplane in mid-flight, with the sun's rays shining off its metal body.",\\
                    & "A satellite photo of an airplane would show a plane in flight, with its wings outstretched and its body elongated.",\\
                    & "The photo would show a plane in the sky, with a trail of clouds behind it.",\\
                    & "A satellite photo of an airplane might show the plane's shadow on the ground, the contrail of the plane's exhaust, or the plane itself flying through the air.",\\
                    & "A satellite photo of an airplane might show the plane in flight, with its contrails stretched out behind it.",\\
                    & "At an airport, a satellite photo would show planes taxiing on the runway and waiting to take off.",\\
                    & "An airplane satellite photo would show a small, fast-moving object with wings and a tail.",\\
                    & "An airplane satellite photo would show a large metal object with wings and propellers flying through the sky.",\\
                    & "A satellite photo of an airplane shows a large metal object with wings and a tail, flying through the air."\\
            \hline
        \end{tabular}
        }
        \caption{10 sample CuPL prompts from each dataset}
        \label{tab:sample_cupl_prompts}
    \end{center}
\end{table}

\chapter{D3G Appendix}
\label{appendix:d3g_appendix}
\section{Additional Prompts}
\begin{table}[h]
    \begin{center}
        \begin{tabular}{|c|c|c|}
            \hline
            \textbf{Demographic} & \textbf{Prompt} & \textbf{Text}\\
            \hline
            Race 7 & "A photo of a $<$race$>$ person" & A photo of a black person\\
            \hline
            Race 4 & "A photo of a $<$race$>$ person" & A photo of a black person\\
            \hline
            Profession & "A photo of a $<$race$>$ $<$prof$>$" & A photo of a black doctor\\
            \hline
            Gender & "A photo of a $<$race$>$ $<$gender$>$" & A photo of a black male\\
            \hline
            Age & "A photo of a $<$age$>$ year old $<$race$>$ person" & A photo of a 30-39 year old black person\\
            \hline
        \end{tabular}
        \caption[Example diverse demographic texts for classifying race 7]{Example diverse demographic texts for classifying \textbf{\textit{race 7}}. The texts for classifying race 4 will be identical, just fewer racial classes.}
        \label{tab:d3g_race_prompts}
    \end{center}
\end{table}
\begin{table}[h]
    \begin{center}
        \begin{tabular}{|c|c|c|}
            \hline
            \textbf{Demographic} & \textbf{Prompt} & \textbf{Text}\\
            \hline
            Gender & "A photo of a $<$gender$>$" & A photo of a female\\
            \hline
            Profession & "A photo of a $<$gender$>$ doctor" & A photo of a female doctor\\
            \hline
            Race 7 & "A photo of a $<$race$>$ $<$gender$>$" & A photo of a black female\\
            \hline
            Race 4 & "A photo of a $<$race$>$ $<$gender$>$" & A photo of a black female\\
            \hline
            Age & "A photo of a $<$age$>$ year old $<$gender$>$ person" & A photo of a 30-39 year old female\\
            \hline
        \end{tabular}
        \caption[Example diverse demographic texts for classifying gender]{Example diverse demographic texts for classifying \textbf{\textit{gender}}.}
        \label{tab:d3g_gender_prompts}
    \end{center}
\end{table}
\begin{table}[h]
    \begin{center}
        \scalebox{0.95}{
        \begin{tabular}{|c|c|c|}
            \hline
            \textbf{Demographic} & \textbf{Prompt} & \textbf{Text}\\
            \hline
            Age & "A photo of a $<$age$>$ year old" & A photo of a 30-39 year old\\
            \hline
            Profession & "A photo of a $<$age$>$ year old doctor" & A photo of a 30-39 year old doctor\\
            \hline
            Race 7 & "A photo of a $<$age$>$ year old $<$race$>$ person" & A photo of a 30-39 year old black person\\
            \hline
            Race 4 & "A photo of a $<$age$>$ $<$race$>$" & A photo of a 30-39 year old black person\\
            \hline
            Gender & "A photo of a $<$age$>$ $<$gender$>$" & A photo of a 30-39 year old female\\
            \hline
        \end{tabular}
        }
        \caption[Example diverse demographic texts for classifying gender]{Example diverse demographic texts for classifying \textbf{\textit{gender}}.}
        \label{tab:d3g_age_prompts}
    \end{center}
\end{table}
\begin{table}[h]
    \begin{center}
        \scalebox{0.95}{
        \begin{tabular}{c|c|c|c|c|c|c|c|c|}
            \hline
               & \textbf{Prompt} & \textbf{White} & \textbf{Black} & \textbf{Latino} & \textbf{East Asian} & \textbf{South East Asian} & \textbf{Indian} & \textbf{Middle Eastern}\\
            \hline
            \multirow{5}{*}{\rotatebox{90}{Race 7}} & Profession & 68.92 & 70.90 & 15.38 & 43.46 & 20.59 & 57.58 & 13.24\\
            \cline{2-9}
            & Race 7 & 9.40 & 61.94 & 15.38 & 66.24 & 5.88 & 60.61 & 26.48\\
            \cline{2-9}
            & Race 4 & - & - & - & - & - & - & -\\
            \cline{2-9}
            & Gender & 20.72 & 67.91 & 15.38 & 44.73 & 11.76 & 51.52 & 29.86\\
            \cline{2-9}
            & Age & 11.57 & 65.67 & 11.54 & 59.92 & 14.71 & 69.70 & 25.07\\
            \hline
        \end{tabular}
        }
        \caption[Average D3G per-class results when classifying the specified demographic]{\textbf{\textit{Average D3G}} per-class results when classifying the specified demographic. Note that all of the prompts are as described in \cref{tab:d3g_prompts}.}
        \label{tab:d3g_avg_race_per-class}
    \end{center}
\end{table}

\section{Additional Per-Class Results}
\begin{table}[h]
    \begin{center}
        \scalebox{0.85}{
        \begin{tabular}{c|c|c|c|c|c|c|c|c|c|c|c|}
            \hline
              & Prompt & \textbf{Chef} & \textbf{Doctor} & \textbf{Engineer} & \textbf{Farmer} & \textbf{Firefighter} & \textbf{Judge} & \textbf{Mechanic} & \textbf{Pilot} & \textbf{Police} & \textbf{Waiter}\\
            \hline
            \multirow{5}{*}{\rotatebox{90}{Profession}} & Profession & 95.54 & 98.80 & 97.62 & 98.36 & 94.12 & 98.20 & 86.49 & 99.37 & 100.0 & 84.34\\
            \cline{2-12}
            & Race 7 & 94.27 & 98.80 & 96.43 & 98.36 & 96.08 & 99.40 & 81.08 & 98.74 & 100.0 & 84.94\\
            \cline{2-12}
            & Race 4 & 94.90 & 98.80 & 96.43 & 98.36 & 96.08 & 99.40 & 82.43 & 98.74 & 100.0 & 84.34\\
            \cline{2-12}
            & Gender & 96.82 & 98.80 & 96.43 & 96.72 & 98.04 & 99.40 & 90.54 & 99.37 & 100.0 &  87.35\\
            \cline{2-12}
            & Age & 97.45 & 98.19 & 96.43 & 98.36 & 98.04 & 98.80 & 85.14 & 97.48 & 100.0 & 80.72\\
            \hline
        \end{tabular}
        }
        \caption[Standard D3G per-class results when classifying the specified demographic]{\textbf{\textit{Standard D3G}} per-class results when classifying the specified demographic. Note that all of the prompts are as described in \cref{tab:d3g_prompts}. These results for race 7 are shown in \cref{tab:d3g_race_per-class}.}
        \label{tab:d3g_prof_per-class}
    \end{center}
\end{table}
\begin{table}[h]
    \begin{center}
        \scalebox{0.85}{
        \begin{tabular}{c|c|c|c|c|c|c|c|c|c|c|c|}
            \hline
              & Prompt & \textbf{Chef} & \textbf{Doctor} & \textbf{Engineer} & \textbf{Farmer} & \textbf{Firefighter} & \textbf{Judge} & \textbf{Mechanic} & \textbf{Pilot} & \textbf{Police} & \textbf{Waiter}\\
            \hline
            \multirow{5}{*}{\rotatebox{90}{Profession}} & Profession & 93.63 & 98.80 & 98.81 & 100.0 & 94.12 & 96.41 & 83.78 & 99.37 & 100.0 & 90.36\\
            \cline{2-12}
            & Race 7 & 93.63 & 98.80 & 97.62 & 98.36 & 94.12 & 97.60 & 81.08 & 98.74 & 100.0 & 90.36\\
            \cline{2-12}
            & Race 4 & 93.63 & 98.80 & 96.43 & 98.36 & 94.11 & 98.80 & 82.43 & 98.74 & 100.0 & 87.35\\
            \cline{2-12}
            & Gender & 96.18 & 98.80 & 96.43 & 98.36 & 98.04 & 99.40 & 90.54 & 99.37 & 100.0 & 89.16\\
            \cline{2-12}
            & Age & 94.90 & 98.80 & 95.24 & 100.0 & 94.12 & 96.41 & 82.43 & 95.60 & 100.0 & 92.77\\
            \hline
        \end{tabular}
        }
        \caption[Average D3G per-class results when classifying the specified demographic]{\textbf{\textit{Average D3G}} per-class results when classifying the specified demographic. Note that all of the prompts are as described in \cref{tab:d3g_prompts}}
        \label{tab:d3g_avg_prof_per-class}
    \end{center}
\end{table}
\end{document}